\documentclass[12pt]{article}
\usepackage{amssymb,amsmath,amsthm,epsfig}

\usepackage{eepic}
\usepackage{epic}
\usepackage{graphicx}
\usepackage{color}
\usepackage{ifpdf}
\usepackage[title]{appendix}

\usepackage{dsfont}
\usepackage{graphicx} 
\usepackage{subfigure}
\usepackage{algorithm}
\usepackage{algorithmic}

\usepackage{amsmath}
\usepackage{bm}
\usepackage{multirow}

\usepackage{caption}

\usepackage{tikz}
\usetikzlibrary{shapes.geometric,arrows}
\usepackage{makecell}

\theoremstyle{plain}
\newtheorem{lem}{Lemma}[section]
\newtheorem{thm}[lem]{Theorem}

\newtheorem{prop}[lem]{Proposition}

\theoremstyle{definition}

\theoremstyle{remark}
\newtheorem{rem}{Remark}[section]

\graphicspath{{images/}}

\newcommand{\bx}{\boldsymbol{x}}
\newcommand{\bX}{\boldsymbol{X}}
\newcommand{\by}{\boldsymbol{y}}
\newcommand{\bY}{\boldsymbol{Y}}
\definecolor{rred}{HTML}{E22626}

\begin{document}
	
	\title{\large\bf {Ensemble Transport Filter via Optimized Maximum Mean Discrepancy}}
	\author{Dengfei Zeng\thanks{School of Mathematical Sciences,  Tongji University, Shanghai 200092, China. (\tt{dfzeng@tongji.edu.cn}).}
		\and
		Lijian Jiang\thanks{School of Mathematical Sciences,  Tongji University, Shanghai 200092, China. (\tt{ljjiang@tongji.edu.cn}), Corresponding author.}
	}
	\date{}
	\maketitle
	\begin{center}{\bf Abstract}
	\end{center}\smallskip
	In this paper, we present a new ensemble-based filter method by reconstructing the analysis step of the particle filter through a transport map, which directly transports  prior particles to posterior particles. 
	The transport map is constructed through an optimization problem described by the Maximum Mean Discrepancy loss function, which matches the expectation information of the approximated posterior and reference posterior. 
	The proposed method inherits the accurate estimation of the posterior distribution from particle filtering while gives an extension to high dimensional assimilation problems. 
	{To improve the robustness of Maximum Mean Discrepancy, a variance penalty term is used to guide the optimization. It prioritizes minimizing the discrepancy between the expectations of highly informative statistics for the reference posteriors. }
	The penalty term significantly enhances the robustness of the proposed method and leads to a better approximation of the posterior. 
	A few numerical examples are presented to illustrate the advantage of the proposed method over ensemble Kalman filter. 
	
	\smallskip
	
	{\bf keywords:}
	Ensemble Assimilation method, Transport Map, Maximum Mean Discrepancy.

	\section{Introduction}
	Data assimilation aims to obtain a reliable state estimate of the dynamical system by best combining observations and mathematical models \cite{lawDataAssimilationMathematical2015,liuPerronFrobeniusOperator2024,pulidoSequentialMonteCarlo2019}. 
	This approach has evolved into a powerful tool with applications across various fields, including geophysics, atmosphere science, and numerical weather prediction \cite{brajardCombiningDataAssimilation2021,farchiUsingMachineLearning2021,jiangCorrectingNoisyDynamic2022,zhangCoupledDataAssimilation2020}. 
	A key category within sequential data assimilation techniques is the filtering method.
	In the Bayesian framework, filtering consists of two primary steps: the prediction step and the analysis step. The prediction step provides prior information by simulating the transition model numerically, while the analysis step applies Bayes' rule to solve a Bayesian inverse problem, incorporating observational data to refine the state estimate.
	
	The main challenge of the filtering method lies in the analysis step since it is difficult to accurately estimate the posterior distribution when the dynamical system is characterized by a high dimensional state space \cite{chattopadhyayDeepLearningenhancedEnsemblebased2023,zhuImplicitEqualWeights2016}, strong nonlinearly transition and observation operator \cite{kawabataNonGaussianProbabilityDensities2020,majdaFilteringComplexTurbulent2012}, and sparse observations \cite{spantiniCouplingTechniquesNonlinear2022}. 
	A theoretical analysis of optimal nonlinear filtering is investigated  in \cite{liptserStatisticsRandomProcesses2001, liptserStatisticsRandomProcesses1977}. 
	The optimal mean square estimate in filtering is equivalent to the posterior mean. 
	However, it is difficult to determine the exact posterior without particular assumptions on the structure of the filter problem. 
	The ensemble Kalman filter (EnKF) is the most popular data assimilation method for nonlinear systems, which gives a best linear unbias state estimation (BLUE) for weakly nonlinear systems based on a linear transformation \cite{evensenEnsembleKalmanFilter2003,evensenSamplingStrategiesSquare2004,evensenEnsembleKalmanFilter2009}. 
	However, the EnKF ignores the inconsistency of the sensitivity of each particle under a strongly nonlinear system \cite{sakovIterativeEnKFStrongly2012}.
	Moreover, it can not correctly approximate the non-Gaussian prior distribution of the state, which significantly limits the application of the EnKF \cite{mandelConvergenceEnsembleKalman2011}. 
	Another well-known filtering method is particle filtering (PF), which is a Monte-Carlo method based on sequential importance sampling. 
	In the asymptotic sense, the posterior approximated by PF converges to the true posterior of the data assimilation problem as the number of particles increases to infinity \cite{farchiReviewArticleComparison2018,lawDataAssimilationMathematical2015}. 
	However, PF suffers from the curse of dimensionality, which often manifests as particle collapse in sequence updating. 
	The number of particles required in PF grows exponentially with respect to  the dimension of  state space  \cite{carrassiDataAssimilationGeosciences2018}. 
	Resampling can alleviate the challenge  by resetting the particle weights to be the same,  but it is often not enough to offset this shortcoming \cite{snyderObstaclesHighdimensionalParticle2008}. In high-dimensional systems, minor differences in states can result in particle weights differing by several orders of magnitude, and resampling may cause the ensemble to collapse into just a few particles. Local particle filtering methods incorporate the concept of localization into particle filtering, which reduces the risk of particle collapse  \cite{rebeschiniCanLocalParticle2015,vanleeuwenParticleFilteringGeophysical2009}. 
	However, it is challenging to effectively and consistently merge the locally updated particles across domains.
	
	Although Bayesian filtering methods vary in form, they all define explicit or implicit transport maps that propagate the prior distribution to the posterior distribution, based on different assumptions regarding the prior distribution and the likelihood function. 
	The idea of transferring probability measures through a transport map has been well-developed and widely used in fluid dynamics, economics, statistics, machine learning, and many other fields \cite{elmoselhyBayesianInferenceOptimal2012,mongeMemoireTheorieDeblais1781,villaniOptimalTransportOld2009}. 
	In recent years, the study of transport  has gained much  attention to treat the challenges of data assimilation in high-dimensional and non-Gaussian nonlinear systems. 
	Some suitable transport maps can been constructed by optimizing the KL divergence between the distribution of the prior and the distribution corresponding to the inverse transport map of the posterior. 
	Spantini et al., \cite{spantiniCouplingTechniquesNonlinear2022}, proposed an ensemble filtering method based on the coupling technique and obtained a non-Gaussian extension of the ensemble Kalman filter by constructing the Knothe-Rosenblatt rearrangement. 
	{Likelihood-free algorithms are proposed by leveraging a block-triangular structure of transport maps and samples from the joint distribution \cite{pmlr-v235-al-jarrah24a, wangEfficientNeuralNetwork2025}.} 
	Pulido et al., \cite{pulidoSequentialMonteCarlo2019}, construct a sequence of kernel embedded mappings that represent a gradient flow based on the principles of local optimal transport. 
	Hoang et al., \cite{hoangMachineLearningbasedConditional2023}, incorporated machine learning into the conditional mean filter and got a nonlinear transformation based on the condition mean's orthogonal projection property.
	
	In this paper, we propose a transport map that directly transforms the prior particles to the posterior particles by reconstructing the analysis step of PF. 
	We formulate the construction of the transport map as an optimization problem, minimizing the Maximum Mean Discrepancy (MMD) loss function, which matches the moment information of the approximated posterior generated by the transport map with the posterior of the PF.
	In this way, the weight information of the analysis step of PF is converted into updates for the particle states. The proposed method is called the ensemble transport filter.
	In this approach, we define the function space of MMD as a Reproducing Kernel Hilbert Space (RKHS), allowing the MMD to be efficiently computed using a kernel function. 
	When the kernel function is a universal kernel, it guarantees that the distribution obtained via the transport map converges to the reference posterior provided by the particle filter as the MMD loss approaches zero.
	{To improve the robustness of MMD in handling high-dimensional state spaces \cite{ramdasDecreasingPowerKernel2015}, we introduce a penalty term based on the variances of the reference distributions.} This encourages the method to prioritize minimizing the mean discrepancy for highly informative statistics. 
	The variance penalty term significantly enhances the method’s robustness in approximating the filtering posterior, making it more effective and stable.  
	Numerical examples demonstrate that the ensemble transport filter with a variance penalty term performs robustly in high-dimensional and non-Gaussian problems. 
	For high-dimensional cases, we locally split the observation into multiple lower-dimensional components. The transport map for the original problem is then constructed by composing the transport maps derived from the separate Bayesian inversions. 
	
	The rest of the article is organized as follows. In {Section \ref{sec:2}}, we introduce the nonlinear filtering problem of a stochastic dynamical system and build a connection with  the transfer of distributions. The transport map is constructed in {Section \ref{sec:3}} by optimizing the mean discrepancy between the approximated posterior and reference posterior. To improve the robustness of MMD, a variance penalty term is introduced. Additionally, an extension for high-dimensional problems based on observation splitting is discussed. Numerical examples are presented to evaluate the performance of the ensemble transport filter in {Section \ref{sec:4}}. Finally, we briefly conclude this article in {Section \ref{sec:5}}.
	
	\section{Bayesian formulation of nonlinear filter}\label{sec:2}
	This section reviews the nonlinear filtering problem of stochastic dynamical systems. Then we reformulate   the filtering process as a transfer from prior distribution to posterior distribution.
	
	\subsection{State estimation of dynamical system}
	Consider the continuous-time dynamical model
	\begin{equation*}
		\mathrm{d}\bX_t = \mathbf{M}(\bX_t) \mathrm{d}t + \mathrm{d} \mathbf{N}_{\bx} \label{eq:filter-prob},
	\end{equation*}
	where the model noise $\mathbf{N}_{\bx}$ is a $\mathbb{R}^n$-valued stochastic process. Most often, $\mathbf{N}_{\bx}$ is set to be a Brownian motion, but sometimes Levy processes are used to account for non-Gaussian noise \cite{spantiniCouplingTechniquesNonlinear2022}. The system is  discretized in time for simulation. Suppose the discrete stepsize of time is $\Delta t$ , we get $\bX_{(k)\Delta t} = \bX_{(k-1)\Delta t} + \mathbf{M}(\bX_{(k-1)\Delta t}) \Delta t + \Delta \mathbf{N}_{\bx}$ with Euler-Maruyama scheme\cite{higham.AlgorithmicIntroductionNumerical2001}. This leads to  a discrete stochastic dynamical model,
	\begin{equation}\label{eq:dynamical model}
		\bX_{k} = \mathbf{M}_{k}(\bX_{k-1}) + \boldsymbol{\eta}_{k},
	\end{equation}
	where $\mathbf{M}_k$ is the transition  model and  $\boldsymbol{\eta}_k$, with $k=\{0, 1, 2,\cdots\}$, are the model noise that is assumed to be statistical independent. {Noisy observations of state $\bX_k$ are available at discrete times
		\begin{equation}\label{eq:observation operator}
			\bY_{k} = \mathbf{H}_{k}(\bX_k) + \boldsymbol{\epsilon}_{k}.
		\end{equation}
		where $\mathbf{H}_k: \mathbb{R}^m\to \mathbb{R}^n$ is the observation operator and $\boldsymbol{\epsilon}_{k}$ is the observation noise. }
	
\begin{figure}[!h]
		\centering
		\includegraphics[width=.6\textwidth]{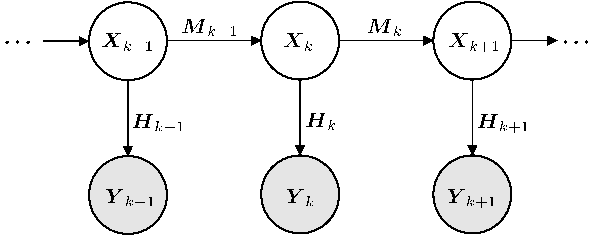}
		\caption{Directed probability graph of dynamical model \eqref{eq:dynamical model} and observation operator \eqref{eq:observation operator} in the form of state-space model}
		\label{fig:ssm}
	\end{figure}

	The discrete representation of dynamical model \eqref{eq:dynamical model} and observation operator \eqref{eq:observation operator} can be seen as a state-space model.  Figure \ref{fig:ssm} shows the directed probability graph of state space model. With the assumed random nature of both model and observation noise, the model states and observations can be described as random variables. Filtering aims to infer the conditional posterior $\pi_{\bX_{k}\mid \bY_{1:k}}$ sequentially, and then a Bayesian framework can be used. 
	
	\subsection{Sequential Bayesian Filtering}\label{sec:2.2}
	{Define the sequence of model states and observations by  $\bx_{0:K}=\{\bx_0, \bx_1,\cdots, \bx_K\}$ and $\by_{1:K}=\{\by_1,\by_2,\cdots, \by_K\}$,  respectively. The aim of sequential Bayesian filtering is to estimate the $\pi_{\bX_k\mid \bY_{1:k}}$ iteratively.} Assuming the prior of the initial state $\bX_0$ to be $\pi_{\bX_0}$, the sequential Bayesian filtering of the state-space model consists of two steps: forcast and analysis. 
	
\begin{figure}[!htbp]
		\centering
		\includegraphics[width=.6\textwidth]{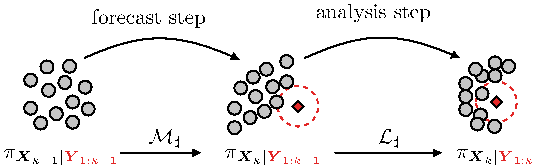}
		\caption{Propagation of particles between distributions in ensemble-based filtering}
		\label{fig:filter}
	\end{figure}

	\begin{itemize}
		\item {\bfseries Forecast}: estimate the prior distribution of iteration step $k$ with the Chapman-Kolmogorov equation
		\begin{equation}\label{eq:C-K equation}
			\pi_{\bX_{k}\mid \bY_{1:k-1}} = \int \pi_{\bX_{k}\mid \bX_{k-1}}\pi_{\bX_{k-1}\mid \bY_{1:k-1}}\mathrm{d} \bX_{k-1},
		\end{equation}
		where $\pi_{\bX_{k}\mid \bX_{k-1}}$ is the transition probability. 
		\item {\bfseries Analysis}: update the conditional posterior distribution at iteration step $k$ based on Bayes' rule
		\begin{equation}\label{eq:bayes rule}
			\pi_{\bX_{k}\mid \bY_{1:k}} = \frac{\pi_{\bX_{k}\mid \bY_{1:k-1}}\pi_{\bY_{k}\mid \bX_{k}}}{\pi_{\bY_{k}\mid \bY_{1:k-1}}},
		\end{equation}
		where $\pi_{\bY_{k}\mid \bX_{k}}$ is the observation likelihood and $\pi_{\bY_{k}\mid \bY_{1:k-1}}$  is the marginal likelihood given by
		\[
		\pi_{\bY_{k}\mid \bY_{1:k-1}}=\int \pi_{\bX_{k}\mid \bY_{1:k-1}}\pi_{\bY_{k}\mid \bX_{k}} \mathrm{d} \bX_{k}.
		\]
	\end{itemize} 
	The integral of Chapman-Kolmogorov equation \eqref{eq:C-K equation} and marginal likelihood $\pi_{\bY_{k}\mid \bY_{1:k-1}}$ of \eqref{eq:bayes rule} are generally intractable to compute analytically under most settings. To overcome this difficulty, ensemble-based filters use an ensemble of particles to represent the uncertainty of states and observations. In ensemble-based filters, the filtering process involves transporting particles between three related probability distributions (see Figure \ref{fig:filter}). 
	
	In the forecast step, the transition density $\pi_{\bX_{k}\mid \bX_{k-1}}$ is defined by a dynamical model, such as \eqref{eq:dynamical model}. 
	Using an ensemble of particles $\{\bx_{k-1}^{(i)}\}_{i=1}^N$ drawn from posterior distribution $\pi_{\bX_{k-1}\mid \bY_{1:k-1}}$ of the previous assimilation window, samples can be directly drawn from transition density $\pi_{\bX_{k}\mid \bX_{k-1}}$ using the discrete dynamical model \eqref{eq:dynamical model}. 
	The resulting particles $\{\hat{\bx}_{k}^{(i)}\}_{i=1}^N$ represent samples from prior distribution $\pi_{\bX_{k}\mid \bY_{1:k-1}}$ of the current assimilation window. 
	In the analysis step, the primary goal is to approximate the posterior distribution at the current assimilation window using the likelihood function and the prior distribution, which is represented by the ensemble of particles. 
	EnKF and its variants approximate the posterior distribution with an ensemble of particles generated from a linear transformation of prior particles under a Gaussian ansatz \cite{sakovIterativeEnKFStrongly2012,spantiniCouplingTechniquesNonlinear2022}.  
	Particle filter adjusts the weights of prior particles through the likelihood function to obtain an accurate estimate of the posterior distribution without altering the state of the prior particles, and a resampling procedure is used to normalize weights \cite{carrassiDataAssimilationGeosciences2018}. 
	
	{For clarity, we denote $\hat{\bX}_k$ as the random variable of forecast distribution $\pi_{{\bX}_{k}\mid \bY_{1:k-1}}$.} From the perspective of transport, the ensemble-based filter involves two distribution transport processes, represented as push-forward operators $\mathcal{M}_{\sharp}$ and $\mathcal{L}_{\sharp}$, as illustrated in Figure \ref{fig:filter}. 
	The  $\mathcal{M}_{\sharp}$ and $\mathcal{L}_{\sharp}$ define the transition probability $\pi_{{\bX}_k\mid \bX_{k-1}}$ and $\pi_{\bX_k\mid \hat{\bX}_k, \bY_{k}}$, respectively,
	\begin{align}
		&\pi_{\hat{\bX}_{k}} = \int \pi_{\hat{\bX}_k\mid \bX_{k-1}}\pi_{\bX_{k-1}\mid \bY_{1:k-1}}\mathrm{d}\bX_{k-1} = \mathcal{M}_{\sharp}\pi_{\bX_{k-1}\mid \bY_{1:k-1}},\label{eq:push-forward dynamic}\\
		&\pi_{\bX_{k}\mid \bY_{1:k}} = \int \pi_{\bX_k\mid \hat{\bX}_k, \bY_{k}}\pi_{\hat{\bX}_{k}}\mathrm{d}\hat{\bX}_{k} = \mathcal{L}_{\sharp}\pi_{\hat{\bX}_{k}}.\label{eq:push-forward observation}
	\end{align}
	Different ensemble filtering methods provide approximate alternatives to the transition probability for the analysis step through different transport maps of ensemble particles. 
	EnKF explicitly constructs a linear transport map, which is efficient and robust in linear and weakly nonlinear systems but performs poorly in strongly nonlinear systems.
	On the other hand, the particle filter can be viewed as an implicit transport map, providing exact filtering but susceptible to particle degeneracy, thus suffering from the curse of dimensionality. 
	In this article, we aim to obtain an approximation of transition probabilities $\pi_{\bX_k\mid \hat{\bX}_k, \bY_{k}}$ by constructing an explicit transport map that reconstructs the analysis step of the particle filter.

	\section{Ensemble Transport Filter}\label{sec:3}
	In this section, we construct a transport map to approximate the transition probability for the analysis step. Section \ref{sec:3.1} formulates this construction as an optimization problem based on the {Maximum Mean Discrepancy} loss by matching the posterior expectation.
	The algorithm for the ensemble transport filter is detailed in Section \ref{sec:3.2}. To enhance the robustness of MMD, a variance penalty is introduced, and then an extension for high-dimensional problems is discussed. Finally, a linear version of the transport filter is derived, which takes a form similar to that of the {Ensemble Kalman Filter}.
	
	\subsection{Proposal transition generated by transport map in analysis step}\label{sec:3.1}
	In filtering, we typically use the posterior mean as the state estimate. Here, we start by considering the posterior expectation of an arbitrary function $f \in \mathcal{C}(\mathbb{R}^n)$. For simplicity, we suppress  the subscripts of the probability density functions in the derivation.
	\begin{equation}\label{eq:post-markov-mean}
		\begin{aligned}
			\mathbb{E}_{\pi_{\bX_{k}\mid \bY_{1:k}}}\left[f(\bx_k)\right] &= \int f(\bx_k)\pi(\bx_k\mid \by_{1:k})\mathrm{d}\bx_k\\
			&=\frac{1}{A}\int f(\bx_k)\pi(\by_k\mid \bx_k)\pi(\bx_k\mid \by_{1:k-1})\mathrm{d}\bx_k\\
			&=\frac{1}{A}\int f(\bx_k)\pi(\by_k\mid \bx_k)\pi(\bx_k\mid \bx_{k-1})\pi(\bx_{k-1}\mid \by_{1:k-1})\mathrm{d}\bx_k\mathrm{d}\bx_{k-1},
		\end{aligned}
	\end{equation}
	in which Bayes' theorem and Chapman-Kolomogrov equation are used, {and $A=\int \pi(\by_k\mid \bx_k)\pi(\bx_k\mid \by_{1:k-1}) \mathrm{d}\bx_k$ is a normalization factor.} 
	
	
	Combining \eqref{eq:push-forward dynamic}-\eqref{eq:push-forward observation},  the posterior expectation \eqref{eq:post-markov-mean} can also be represented by
	\begin{equation*}
		\mathbb{E}_{\pi_{\bX_{k}\mid \bY_{1:k}}}\left[f(\bx_k)\right] = \int f(\bx_k)\pi(\bx_k\mid \hat{\bx}_{k}, \by_k)\pi(\hat{\bx}_k\mid \bx_{k-1})\pi(\bx_{k-1}\mid \by_{1:k-1})\mathrm{d}\hat{\bx}_{k}\mathrm{d}\bx_{k-1}\mathrm{d}\bx_{k}.
	\end{equation*}
	As discussed in \ref{sec:2.2}, the transition probability density $\pi_{\bX_k\mid \hat{\bX}_k, \bY_{k}}$ is implicitly defined through the likelihood function. Here, we aim to find a transition probability density $q^{\mathcal{T}_k}_{\bX_k\mid \hat{\bX}_{k}, \by_{k}}$ through a parametric transport map $\mathcal{T}_k$ to obtain particles from the posterior distribution directly. To eliminate the intermediate variable $\hat{\bX}_{k}$, we denote 
	$$
	q^{\mathcal{T}_k}(\bx_k\mid \bx_{k-1},\by_k) = \int q^{\mathcal{T}_k}(\bx_k\mid \hat{\bx}_{k}, \by_k)\pi(\hat{\bx}_k\mid \bx_{k-1})\mathrm{d}\hat{\bx}_{k}.
	$$
	Then we have 
	$$
	\mathbb{E}_{\pi_{\bX_{k}\mid \bY_{1:k}}}\left[f(\bx_k)\right]=\int f(\bx_k)q^{\mathcal{T}_k}(\bx_k\mid \bx_{k-1},\by_k)\pi(\bx_{k-1}\mid \by_{1:k-1})\mathrm{d}\bx_k\mathrm{d}\bx_{k-1}.
	$$ 
	Moreover, insert $q^{\mathcal{T}_k}(\bx_k\mid \bx_{k-1},\by_k)$ into \eqref{eq:post-markov-mean}, we can rewrite it as
	\begin{equation}
		\begin{aligned}
			\mathbb{E}_{\pi_{\bX_{k}\mid \bY_{1:k}}}\left[f(\bx_k)\right] &=\\
			\int f(\bx_k)&h^{\mathcal{T}_k}(\bx_{k-1:k}, \by_k)q^{\mathcal{T}_k}(\bx_k\mid \bx_{k-1},\by_k)\pi(\bx_{k-1}\mid \by_{1:k-1})\mathrm{d}\bx_k\mathrm{d}\bx_{k-1},
		\end{aligned}
	\end{equation}
	where $h^{\mathcal{T}_k}(\bx_{k-1:k}, \by_k) = \frac{\pi(\by_{k}\mid \bx_{k})\pi(\bx_k\mid \bx_{k-1})}{Aq^{\mathcal{T}_k}(\bx_k\mid \bx_{k-1},\by_k)}$. 
	
	If a suitable transport map $\mathcal{T}_k$ has been chosen, the transition probability density $q^{\mathcal{T}_k}_{\bX_k\mid \hat{\bX}_{k}, \by_{k}}$ corresponding to the transport map should satisfy
	\begin{equation}\label{eq:equiv condition}
		h^{\mathcal{T}_k}(\bx_{k-1:k}, \by_k) \equiv 1.
	\end{equation}
	Directly solving \eqref{eq:equiv condition} is challenging, as the normalization constant $A$  is generally inaccessible. Furthermore, an analytical form for the transition probability corresponding to the nonlinear transport map cannot be derived. However, the transport map $\mathcal{T}_k$ can be computed indirectly.
	
	In a compact form, we can use the Chapman-Kolmogorov formula to obtain an transported posterior probability density, 
	$$
	q^{\mathcal{T}_k}(\bx_k\mid \by_{1:k}) = \int q^{\mathcal{T}_k}(\bx_k\mid \bx_{k-1},\by_k)\pi(\bx_{k-1}\mid \by_{1:k-1})\mathrm{d}\bx_{k-1}.
	$$
	Next, we construct the desired transport map indirectly by matching posterior expectations. To begin, we introduce the following lemma.
	
	\begin{lem}\label{lm:1}(Lemma 9.3.2 in \cite{richardmRealAnalysisProbability2018})
		Let $(\Omega, d)$ be a metric space, and $p, q$ be two probability measure defined on $\Omega$. Then $p=q$ if and only if $\mathbb{E}_{\bx\sim p}[f(\bx)] = \mathbb{E}_{\by\sim q}[f(\by)]$ for all $f\in \mathcal{C}(\Omega)$, where $\mathcal{C}(\Omega)$ is the space of bounded continuous on $\Omega$.
	\end{lem}
	The following proposition is a direct result of Lemma \ref{lm:1}, and it suggests that the transported posterior $q^{\mathcal{T}_k}(\bx_k\mid \by_{1:k})$ gives an approximation of the true posterior probability density under certain conditions.
	
	\begin{prop}\label{prop:1}
		Given transport map $\mathcal{T}_k: \mathbb{R}^n\to \mathbb{R}^n$, if the expectation of posterior and true posterior are matched for all $\ f \in \mathcal{C}(\mathbb{R}^n)$,
		$$
		\int f(\bx_k)\pi(\bx_{k}\mid \by_{1:k})\mathrm{d}\bx_{k}=\int f(\bx_k)q^{\mathcal{T}_k}(\bx_{k}\mid \by_{1:k})\mathrm{d}\bx_{k},
		$$
		then the approximated posterior $q^{\mathcal{T}_k}(\bx_k\mid \by_{1:k})$ is identical to the true posterior and $\mathcal{T}_k$ is the desired transport map that transfer prior to posterior.
	\end{prop}
	
	Based on Propsition \ref{prop:1}, we can reformulate the problem of finding transport map $\mathcal{T}_k$ as the optimization probem, 
	\begin{equation}\label{eq:opti_prob}
		\mathcal{T}_k = \underset{\mathcal{S}}{\arg \min }\ \underset{ f\in \mathcal{C}(\mathbb{R}^n)}{\sup} \left(\mathbb{E}_{\bx\sim\pi_{\bX_{k}\mid \bY_{1:k}}}[f(\bx)] - \mathbb{E}_{\by\sim\pi^{\mathcal{S}}_{\bX_{k}\mid \bY_{1:k}}}[f(\by)]\right),
	\end{equation}
	{where $\pi^{\mathcal{S}}_{\bX_{k}\mid \bY_{1:k}}$ is the approximated posterior given by transport map $\mathcal{S}$.} It is challenging to solve the optimization  on the continuous function space $\mathcal{C}(\Omega)$.
	To make it efficient and feasible, we select an RKHS space $\mathcal{H}$ defined on a compact metric space $\Omega$ with reproducing kernel $k(\cdot, \cdot)$. 
	{According to the Theorem 3 of \cite{NIPS2006_e9fb2eda}, the properties of the Lemma \ref{lm:1} are held when the kernel of $\mathcal{H}$ is universal. }
	Let $\mathcal{F}$ be a unit ball in RKHS $\mathcal{H}$. Then $\operatorname{MMD}[p, q;\mathcal{F}]=0$ if and only if $p=q$.
	The target of the optimization problem \eqref{eq:opti_prob} can be rewritten as
	\begin{equation}\label{eq:MMD}
		\underset{f\in \mathcal{F}}{\sup} \left(\mathbb{E}_{\bx\sim\pi_{\bX_{k}\mid \bY_{1:k}}}[f(\bx)] - \mathbb{E}_{\by\sim\pi^{\mathcal{S}}_{\bX_{k}\mid \bY_{1:k}}}[f(\by)]\right).\\
	\end{equation}
	{The formula \eqref{eq:MMD} is also called  Maximum Mean Discrepancy}. The MMD is a reasonable measure of the difference between two distributions. When MMD equals zero, two distributions are identical. With the reproducing property of RKHS, we do not need to compute the expectation of a specific function $f$ directly. It suffices to compute \eqref{eq:MMD} by   \eqref{eq:MMD kernel form} in  the following lemma.  
	
	\begin{lem}\label{lem:MMD in kernel form}(Lemma 5 in \cite{NIPS2006_e9fb2eda})
		Given $\bx$ and $\bx'$ independent variables with distribution $p$ , $\by$ and $\by'$ independent variables with distribution $q$. Assuming that the expectation of feature mapping $\phi(\cdot)$ of RKHS $\mathcal{H}$ exists.  Then 
		\begin{equation}\label{eq:MMD kernel form}
			\operatorname{MMD}\limits^2[p, q;\mathcal{F}] = \mathbb{E}_{\bx, \bx'\sim p}[k(\bx, \bx')] - 2\mathbb{E}_{\bx\sim p, \by\sim q}[k(\bx, \by)] + \mathbb{E}_{\by, \by'\sim p}[k(\by, \by')].
		\end{equation}
	\end{lem}

	\subsection{Ensemble approximation of transport filter}\label{sec:3.2}
	Motivated by the nudging method, we can construct the transport map $\mathcal{T}_k$ by
	\begin{equation}\label{eq:transport}
		\bX_{k} = \mathcal{T}_k(\hat{\bX}_{k}) = \hat{\bX}_{k} + \mathbf{T}(\bY_{k} - \mathbf{H}(\hat{\bX}_{k})),
	\end{equation}
	where $\mathbf{T}(\cdot)$ can be  a linear or nonlinear map. Transport map $\mathcal{T}_k$ generates a transition probability that convert $\pi_{\bX_{k}\mid \bY_{1:k-1}}$ to $q^{\mathcal{T}_k}_{\bX_{k}\mid \bY_{1:k}}$ and is related to the observation $\by_{k}$. 
	We select a parameterized function space $\mathcal{G}_0$ and take map $\mathbf{T}$ to be the function in $\mathcal{G}_0$. Then $\mathcal{G}_0$ is transformed into space $\mathcal{G}$ according to \eqref{eq:transport}. 
	We now aims to find an optimal transport map $\mathcal{T}_k^*\in \mathcal{G}$ that minimizes  the $\text{MMD}^2$ loss
	
	\begin{equation}\label{eq:loss}
		\mathcal{T}_k^* = \underset{\mathcal{T}_k\in \mathcal{G}}{\arg \min} \operatorname{MMD}^2\left[\pi^{\mathcal{T}_k}_{\bX_{k}\mid \bY_{1:k}}, \pi_{\bX_{k}\mid \bY_{1:k}}; \mathcal{F}\right],
	\end{equation}
	where $\mathcal{F}$ is a unit ball of RKHS $\mathcal{H}$. 
	
	In the framework of  ensemble-based filter method, we start from an ensemble of particles $\{\bx_{k-1}^{(i)}\}_{i=1}^N$ from the posterior density at time step $k-1$. Samples $\{\hat{\bx}_{k}^{(i)}\}_{i=1}^N$ of the prior density at time step $k$ can be generated by the discrete dynamical model $\eqref{eq:dynamical model}$. Then the samples of prior density can be pushed forwards $q^{\mathcal{T}_k}_{\bX_{k}\mid \bY_{1:k}}$ by transport map $\mathcal{T}_k$ in \eqref{eq:transport}. {In particle filtering}, information on the true posterior density is implied  in the weights of particles. We can express the reference posterior density as
	\begin{equation}\label{eq:particle filter posterior}
		\pi_{\bX_{k}\mid \bY_{1:k}} \approx \sum_{i=1}^N w_i \delta_{\hat{\bx}_{k}^{(i)}}(\bx_k ), w_i = \frac{\pi(\by_k\mid \hat{\bx}_{k}^{(i)})}{\sum_{j=1}^N\pi(\by_k\mid \hat{\bx}_{k}^{(j)})}.
	\end{equation}
	The empirical estimation of $\text{MMD}^2$ loss between approximated and true posterior density is given by the following formula \cite{yanWeightedClassSpecificMaximum2020},
	\begin{equation}\label{eq:empirical_MMD}
		\begin{aligned}
			\operatorname{MMD}^2 = \sum_{i,j=1}^{m}&w_iw_jk(\hat{\bx}_{k}^{(i)}, \hat{\bx}_{k}^{(j)})-2\sum_{i,j=1}^{m}w_iv_jk(\hat{\bx}_{k}^{(i)}, \mathcal{T}_k(\hat{\bx}_{k}^{(j)}))\\
			&+\sum_{i,j=1}^{n}v_iv_jk(\mathcal{T}_k(\hat{\bx}_{k}^{(i)}), \mathcal{T}_k(\hat{\bx}_{k}^{(j)})),
		\end{aligned}
	\end{equation}
	where $\{v_i\}_{i=1}^N$ are weights of particles generated by transport map $\mathcal{T}_k$ and $v_i=1/N$. 
	
	\begin{rem}
		The selection of the kernel function in the ensemble transport filter is crucial in determining how well the ensemble transport filter approximates the posterior distribution. When we choose a universal kernel such as { the Gaussian kernel $k(\bx, \bx^{\prime})=\exp\left(-\|\bx-\bx^{\prime}\|/r^2\right)$}, ensemble transport filter will tend to approximate the posterior distribution as accurate as possible. When we choose a linear kernel function $k(\bx, \bx^{\prime})=\bx^{\top}\bx'+1$, it tends to obtain a better estimate of the mean of the posterior distribution and ignores higher order moment information. For further analysis, one can adaptively adjust the kernel functions based on the actual model, as discussed in \cite{wangPhysicsInformedDeep2022}.
	\end{rem}
	\begin{rem}
		A suitably chosen function space $\mathcal{G}$ helps to improve the performance of the ensemble transport filter; one of the simplest choices is to take T to be a linear map as in EnKF, where all particles are uniformly complementary concerning the observation. This approach is computationally efficient.  However, like EnKF, it performs poorly in strongly nonlinear cases. The  function space can be learned by  normalized flow, RBF mapping  and particle flow \cite{spantiniCouplingTechniquesNonlinear2022}. In our numerical simulations,  we  learn  it by  fully connected neural networks.
	\end{rem}
	\begin{algorithm}
		\renewcommand{\algorithmicrequire}{\textbf{Input:}}
		\renewcommand{\algorithmicensure}{\textbf{Output:}}
		\caption{Ensemble Transport map filter(EnTranF)}
		\begin{algorithmic}[1]
			\REQUIRE Given ensemble $\{\bx_{k-1}^{(i)}\}_{i=1}^N$ at previous time step $k-1$, current observation $\by^o_k$, discrete dynamical model $\mathbf{M}_{k}(\cdot)$, observation operator $\mathbf{H}_k(\cdot)$, model uncertainty $\pi(\boldsymbol{\eta}_{k})$ and likelihood function $\pi(\by_k\mid \bx_k)$;
			\FOR {$i=1,\cdots, N$} 
			\STATE $\hat{\bx}_{k}^{(i)} = \mathbf{M}_{k}(\bx_{k-1}^{(i)}) + \boldsymbol{\eta}_{k}^{(i)}$; 
			\ENDFOR \COMMENT{Forecast Step}
			\STATE Calculate particle weights $w_i = \frac{\pi(\by^o_k\mid \hat{\bx}_{k}^{(i)})}{\sum_{j=1}^N\pi(\by^o_k\mid \hat{\bx}_{k}^{(j)})}$;
			\STATE Initialize transport map $\mathcal{T}_k$ defined as \eqref{eq:transport};
			\REPEAT
			\FOR{$i=1,\cdots, N$}
			\STATE ${\bx}_{k}^{(i)} = \mathcal{T}_k(\hat{\bx}_{k}^{(i)})$; \COMMENT{Alternative Analysis Step}
			\STATE $v_i = 1/N$;
			\ENDFOR
			\STATE Calculate $\text{MMD}^2$ loss by \eqref{eq:empirical_MMD};
			\STATE Update transport map $\mathcal{T}_k$ using gradient descent.
			\UNTIL{Stopping criterion met}
			\ENSURE Approximated posterior ensemble $\{{\bx}_{k}^{(i)}\}_{i=1}^N$;
		\end{algorithmic}
	\end{algorithm}

	\subsubsection{Variance penalty for MMD loss}\label{sec:3.2.1}
	Given the nonlinear nature of the optimization problem, the learned transport map may be locally optimal. Additionally, vanilla MMD loss is less efficient in high dimensional problems. {To make the transport map better approximate the posterior distribution, we introduce a variance penalty term. It prioritizes  minimizing the discrepancy between the expectations of highly informative statistics for the reference posteriors. Suppose $p$ is the reference distribution and $q$ the approximation posterior.} Since $\mathcal{F}$ is a unit ball of RKHS, the symmetry implies that
	\[
	\left[\underset{f\in \mathcal{F}}{\sup} \left(\mathbb{E}_{\bx\sim p}[f(\bx)] - \mathbb{E}_{\by\sim q}[f(\by)]\right)\right]^2 = \underset{f\in \mathcal{F}}{\sup} \left(\mathbb{E}_{\bx\sim p}[f(\bx)] - \mathbb{E}_{\by\sim q}[f(\by)]\right)^2.
	\] 
	{The MMD loss with variance penalty is defined by 
		\begin{equation}\label{eq:penalty MMD}
			\text{Loss} = \underset{f\in \mathcal{F}}{\sup} \left\{\left(\mathbb{E}_{\bx\sim p}[f(\bx)] - \mathbb{E}_{\by\sim q}[f(\by)]\right)^2 + 2\mathbb{V}ar_{\bx\sim p}[f(\bx)]\right\}.
		\end{equation}
		The penalty term does not depend on the approximate distribution. As a result, when minimizing the loss function, it does not compromise the ability of the MMD term to enforce accurate distributional matching between the approximated distribution $q$ and the reference $p$.
		By a direct calculation, we can rewrite \eqref{eq:penalty MMD} as 
		\begin{equation}\label{eq:Component of MMD}
			\begin{aligned}
				\operatorname{Loss} &= \underset{f\in \mathcal{F}}{\sup}\left\{\left(\mathbb{E}_{\bx\sim p,\atop \by\sim q}\left[f(\bx)-f(\by)\right]^2\right) + \left(\mathbb{V}ar_{\bx\sim p}[f(\bx)] - \mathbb{V}ar_{\by\sim q}[f(\by)]\right)\right\}\\
				&\le \underset{f\in \mathcal{F}}{\sup}\ \mathbb{E}_{\bx\sim p,\atop \by\sim q}\left[f(\bx)-f(\by)\right]^2 + \underset{f\in \mathcal{F}}{\sup}\ \left(\mathbb{V}ar_{\bx\sim p}[f(\bx)] - \mathbb{V}ar_{\by\sim q}[f(\by)]\right), 
			\end{aligned}
		\end{equation}
		the equality holds when $p=q$. In \eqref{eq:Component of MMD}, training loss is split into two terms.  The first term $\mathbb{E}_{\bx\sim p,\atop \by\sim q}\left[f(\bx)-f(\by)\right]^2$ constrain the expectation of the $L^2$-distance between the samples drawn from two densities to be close, while the second term $\left(\mathbb{V}ar_{\bx\sim p}[f(\bx)] - \mathbb{V}ar_{\by\sim q}[f(\by)]\right)$ forces the discrepancy of the variance of two distribution to be close.} Since $f \in \mathcal{F}$ and $\|f\|_{\mathcal{H}}\le 1$, then
	\[
	\begin{aligned}
		\underset{f\in \mathcal{F}}{\sup} \mathbb{E}\left[f(\bx)-f(\by)\right]^2 &= \underset{f\in \mathcal{F}}{\sup} \mathbb{E}\left[\langle \phi_{\bx}-\phi_{\by}, f\rangle_{\mathcal{H}}^2\right]\\
		&=\underset{f\in \mathcal{F}}{\sup} \mathbb{E}\left[\|\phi_{\bx}-\phi_{\by}\|_{\mathcal{H}}^2\left\langle \frac{\phi_{\bx}-\phi_{\by}}{\|\phi_{\bx}-\phi_{\by}\|_{\mathcal{H}}}, f\right\rangle_{\mathcal{H}}^2\right],
	\end{aligned}
	\]
	where $\phi_{\bx} = k(\bx, \cdot)$ and $\phi_{\by}=k(\by, \cdot)$. With Cauchy inequality, it gives
	\begin{equation*}
		\left\langle \frac{\phi_{\bx}-\phi_{\by}}{\|\phi_{\bx}-\phi_{\by}\|_{\mathcal{H}}}, f\right\rangle_{\mathcal{H}} \le \frac{\|\phi_{\bx}-\phi_{\by}\|_{\mathcal{H}}}{\|\phi_{\bx}-\phi_{\by}\|_{\mathcal{H}}} \|f\|_{\mathcal{H}}\le 1.
	\end{equation*}
	When $f = \frac{\phi_{\bx}-\phi_{\by}}{\|\phi_{\bx}-\phi_{\by}\|_{\mathcal{H}}}\in \mathcal{F}$, the equality holds. Thus
	\begin{equation*}
		\underset{f\in \mathcal{F}}{\sup} \mathbb{E}\left[f(\bx)-f(\by)\right]^2 = \mathbb{E}\left[\|\phi_{\bx}-\phi_{\by}\|_{\mathcal{H}}^2\right].
	\end{equation*}
	Then the first term of \eqref{eq:Component of MMD} can be rewritten as
	\begin{equation}\label{eq:penalty MMD-1}
		\underset{f\in \mathcal{F}}{\sup} \mathbb{E}\left[f(\bx)-f(\by)\right]^2 = \mathbb{E}_{\bx\sim p}[k(\bx, \bx)] - 2\mathbb{E}_{\bx\sim p, \by\sim q}[k(\bx, \by)] + \mathbb{E}_{\by\sim p}[k(\by, \by)].
	\end{equation}
	{The second term is actually the maximum covariance discrepancy (MCD) that proposed in \cite{zhangMaximumMeanCovariance2020}. It can be rewritten as
		\begin{equation}
			\underset{f\in \mathcal{F}}{\sup}\ \left(\mathbb{V}ar_{\bx\sim p}[f(\bx)] - \mathbb{V}ar_{\by\sim q}[f(\by)]\right) = \|\mathbb{C}[p]-\mathbb{C}[q]\|_{\text{HS}},
			\label{eq:mcd}
		\end{equation}
		where $\|\cdot\|_{\text{HS}}$ denote the Hilbert-Schmidt norm and $\mathbb{C}[p]$ is defined as $\mathbb{C}[p]=\mathbb{E}_p[\phi(\bx)\otimes\phi(\bx)]-\mathbb{E}_p[\phi(x)]\otimes\mathbb{E}_p[\phi(x)]$ where $\otimes$ is the tensor product in RKHS $\mathcal{H}$. According to Lemma 2 in \cite{zhangMaximumMeanCovariance2020}, we have
		\begin{equation*}
			\begin{aligned}
				\|\mathbb{C}[p]-\mathbb{C}[q]\|_{\text{HS}}^2 =& \mathbb{E}_{\bx, \bx^{\prime}}[k^2(\bx, \bx^{\prime})] - 2\mathbb{E}_{\bx}\left[\mathbb{E}^2_{\bx^{\prime}}[k(\bx, \bx^{\prime})]\right] + \mathbb{E}^2_{\bx, \bx^{\prime}}[k(\bx, \bx^{\prime})]\\
				&-2\mathbb{E}_{\bx, \by}[k^2(\bx, \by)] + 2\mathbb{E}_{\bx}\left[\mathbb{E}^2_{\by}[k(\bx, \by)]\right] \\
				&+ 2\mathbb{E}_{\by}\left[\mathbb{E}^2_{\bx}[k(\by, \bx)]\right] - 2\mathbb{E}^2_{\bx, \by}[k(\bx, \by)] \\
				&+\mathbb{E}_{\by, \by^{\prime}}[k^2(\by, \by^{\prime})] - 2\mathbb{E}_{\by}\left[\mathbb{E}^2_{\by^{\prime}}[k(\by, \by^{\prime})]\right] + \mathbb{E}^2_{\by, \by^{\prime}}[k(\by, \by^{\prime})].
			\end{aligned}
		\end{equation*}
	}
	
	Similar to \eqref{eq:empirical_MMD}, we can estimate \eqref{eq:penalty MMD-1} with ensemble particles, i.e., 
	\begin{equation*}
		\begin{aligned}
			\underset{f\in \mathcal{F}}{\sup} \mathbb{E}\left[f(\bx)-f(\by)\right]^2 = \sum_{i=1}^{m}&w_ik(\hat{\bx}_{k}^{(i)}, \hat{\bx}_{k}^{(i)})-2\sum_{i,j=1}^{m}w_iv_jk(\hat{\bx}_{k}^{(i)}, \mathcal{T}_k(\hat{\bx}_{k}^{(j)}))\\
			&+\sum_{i}^{n}v_ik(\mathcal{T}_k(\hat{\bx}_{k}^{(i)}), \mathcal{T}_k(\hat{\bx}_{k}^{(i)})),
		\end{aligned}
	\end{equation*}
	{And the second term of Equation \eqref{eq:Component of MMD} can be estimated with
		$$
		\|\mathbb{C}[p]-\mathbb{C}[q]\|_{\text{HS}}^2 = \operatorname{tr}(KWKW),
		$$
		where
		$$
		K = \left[\begin{matrix}
			k(\hat{\bx}^{(1)}, \hat{\bx}^{(1)}) & \cdots & k(\hat{\bx}^{(1)}, \hat{\bx}^{(n)}) & k(\hat{\bx}^{(1)}, \hat{\by}^{(1)}) & \cdots & k(\hat{\bx}^{(1)}, \hat{\by}^{(n)})\\
			\vdots & \ddots & \vdots & \vdots & \ddots & \vdots\\
			k(\hat{\bx}^{(n)}, \hat{\bx}^{(1)}) & \cdots & k(\hat{\bx}^{(n)}, \hat{\bx}^{(n)}) & k(\hat{\bx}^{(n)}, \hat{\by}^{(1)}) & \cdots & k(\hat{\bx}^{(n)}, \hat{\by}^{(n)})\\
			k(\hat{\by}^{(1)}, \hat{\bx}^{(1)}) & \cdots & k(\hat{\by}^{(1)}, \hat{\bx}^{(n)}) & k(\hat{\by}^{(1)}, \hat{\by}^{(1)}) & \cdots & k(\hat{\by}^{(1)}, \hat{\by}^{(n)})\\
			\vdots & \ddots & \vdots & \vdots & \ddots & \vdots\\
			k(\hat{\by}^{(n)}, \hat{\bx}^{(1)}) & \cdots & k(\hat{\by}^{(n)}, \hat{\bx}^{(n)}) & k(\hat{\by}_{k}^{(n)}, \hat{\by}^{(1)}) & \cdots & k(\hat{\by}^{(n)}, \hat{\by}^{(n)})
		\end{matrix}\right],
		$$
		and
		$$
		W = \left[\begin{matrix}
			w_1-w_1^2 & w_1w_2 & \cdots & w_1w_n & 0 & 0 & \cdots & 0 \\
			w_2w_1 & w_2 - w_2^2 & \cdots & w_2w_n & 0 & 0 & \cdots & 0 \\
			\vdots & \vdots & \ddots & \vdots & \vdots & \vdots & \ddots & \vdots \\
			w_nw_1 & w_nw_2 & \cdots & w_n -w_n^2 & 0 & 0 & \cdots & 0 \\
			0 & 0 & \cdots & 0 & v_1-v_1^2 & v_1v_2 & \cdots & v_1v_n\\
			0 & 0 & \cdots & 0 & v_2v_1 & v_2 - v_2^2 & \cdots & v_2v_n\\
			\vdots & \vdots & \ddots & \vdots & \vdots & \vdots & \ddots & \vdots \\
			0 & 0 & \cdots & 0 & v_nv_1 & v_nv_2 & \cdots & v_n -v_n^2
		\end{matrix}\right].
		$$
		Here, $\{\bx^{(i)}\}_{i=1}^n$ are weighted samples from $p$ with weights $\{w_i\}_{i=1}^N$ and $\{\by^{(i)}\}_{i=1}^n$ are weighted samples from $q$ with weights $\{v_i\}_{i=1}^N$. In filtering, we take $p$ as the reference distribution and $q$ as the approximated distribution given by transport map $\mathcal{T}_k$ respectively.
	}
	
	\subsubsection{An extension to high dimensional problems}\label{sec:3.2.2}
	As shown in algorithm 1, the proposed method requires the weighted prior samples to act as the reference posterior. It essentially employs a transport map to replace the resampling step in particle filtering, which enhances the diversity of particles.  { In slightly higher dimensional problems, the proposed method can reduce the rate at which particle degeneracy occurs. However, the curse of dimensioanlity remains.}
	In high dimensional problems, the particle degeneracy issue in particle filtering is related to the dimensionality of the observation vector. High-dimensional observation vectors can cause even small differences in each component to result in several orders of magnitude difference in particle weights \cite{vanleeuwenNonlinearDataAssimilation2015}.  
	
	To address high-dimensional problems, we split the observations into spatially distinct components and apply Bayes' theorem separately to each component. {Sequential processing of batches of observations is a standard technique in data assimilation \cite{houtekamerSequentialEnsembleKalman2001, andersonOptimalFiltering1979}.} { We denote the $i$-th component of observation as $\by_{[i],k} = \mathbf{H}_{[i],k}(\bx_k)+\epsilon_{[i],k}, i=1,\cdots, m$.} 
	\begin{figure}[!htbp]
		\centering
		\subfigure[]{
			\begin{minipage}[t]{0.46\textwidth}
				\centering
				\includegraphics{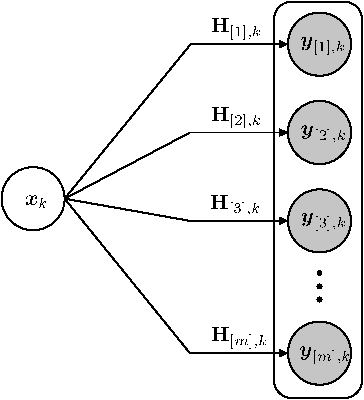}
			\end{minipage}
		}
		\subfigure[]{
			\begin{minipage}[t]{0.46\textwidth}
				\centering
				\includegraphics{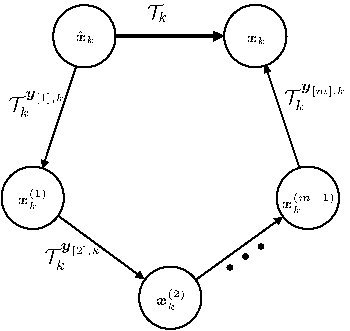}
			\end{minipage}
		}
		\caption{Schematic of observation split and composition of transport maps generated by each splited observation.}
		\label{fig:obsplit}
	\end{figure}
	
	Based on the schematic of observation split shown in panel (a) of Figure \ref{fig:obsplit}, we obtain
	\begin{equation}\label{eq:ob-split joint distro}
			\begin{aligned}
				\pi(\bx_k, \by_k\mid \by_{1:k-1}) &= \pi(\bx_k\mid \by_{1:k-1})\pi(\by_k\mid \bx_k)\\
				&=\pi(\bx_k\mid \by_{1:k-1})\prod_{i=1}^m\pi(\by_{[i],k}\mid \bx_k).
			\end{aligned}
	\end{equation}
	
	We can express \eqref{eq:ob-split joint distro} as a sequence of Bayesian formulas that gradually incorporate observations. 
	
		\begin{equation*}
			\begin{aligned}
				&\pi(\bx_k \mid \by_{1:k-1}, \by_{[1],k}) \propto \pi(\bx_k\mid \by_{1:k-1})\pi(\by_{[1],k}\mid \bx_k),\\
				&\pi(\bx_k \mid \by_{1:k-1}, \by_{[1:2],k}) \propto \pi(\bx_k \mid \by_{1:k-1}, \by_{[1],k})\pi(\by_{[2],k}\mid \bx_k),\\
				&\vdots\\
				&\pi(\bx_k \mid \by_{1:k-1}, \by_{[1:i],k})\propto \pi(\bx_k \mid \by_{1:k-1}, \by_{[1:i-1],k})\pi(\by_{[i],k}\mid \bx_k),\\
				&\vdots\\
				&\pi(\bx_k \mid \by_{1:k-1}, \by_{[1:m],k})\propto \pi(\bx_k \mid \by_{1:k-1}, \by_{[1:m-1],k})\pi(\by_{[m],k}\mid \bx_k),\\
				&\pi(\bx_k \mid \by_{1:k}) = \pi(\bx_k \mid \by_{1:k-1}, \by_{[1:m],k}).
			\end{aligned}
		\end{equation*}
		Let $\bx_{k}^{(i)}=\bx_k \mid \by_{1:k-1}, \by_{[1:i],k}$, then $\pi(\bx_{k}^{(i)})$ is the posterior from the likelihood $\pi(\by_{[i],k}\mid \bx_k^{(i-1)})$ and the prior $\pi(\bx_{k}^{(i-1)})$. Based on the proposed method in Section \ref{sec:3.2.1}, each step in above sequential process can be alternated by a transport map $\mathcal{T}_k^{\by_{[i],k}}$. 
		Then the transport map from filter prior $\pi(\bx_k\mid \by_{1:k-1})$ to filter posterior $\pi(\bx_k\mid \by_{1:k})$ can be presented as a composition of the series of transport map (panel (b) of Figure \ref{fig:obsplit}),
		\begin{equation}
			\mathcal{T}_k = \mathcal{T}_k^{\by_{[m],k}}\circ\mathcal{T}_k^{\by_{[m-1],k}}\circ\cdots\circ\mathcal{T}_k^{\by_{[1],k}}.
		\end{equation}
	
	From the perspective of computational complexity, the proposed observation splitting approach introduces some additional computational costs; However, these are generally acceptable. Firstly, the dimensionality of each individual sub-problem is substantially reduced. Secondly, by converting the problem into multiple low-dimensional observation data assimilation tasks, the required number of particles can be significantly decreased, leading to improved computational efficiency.
	
	\subsubsection{Connection with Ensemble Kalman Filter}
	
	If we  restrict the kernel function to be the linear kernel and the search space of the transport map to be linear, we can obtain an analytical form of the transport map filter. {When using a linear kernel function, the RKHS is the space of linear functions, and MMD only matches the means of the distributions. In this case, we can ignore in \eqref{eq:mcd} and focus solely on the mean matching in \eqref{eq:penalty MMD-1}.}
	
	\begin{thm}\label{thm:EnTranFp LL}
		Given the RKHS kernel $k(\bx, \bx')=\bx^{\top}\bx'+1$. Moreover, we assume \eqref{eq:transport} to be a linear transport, i.e., 
		\begin{equation}\label{eq:linear transport form}
			\bX_{k}^{\mathcal{T}_k} = \hat{\bX}_{k} + \mathbf{T}(\by_{k} + \boldsymbol{\varepsilon}_{k} - \mathbf{H}(\hat{\bX}_{k})),
		\end{equation}
		where $\mathbf{T}\in \mathbb{R}^{m\times n}$, $\by_k=\mathbf{H}_k(\bx_k)$ and $\boldsymbol{\varepsilon}_{k}$ is independent with $\hat{\bX}_{k}$. Denote
		\begin{equation*}
			\begin{aligned}
				\bar{\bX}_k & = \hat{\bX}_{k}-\mathbb{E}[\bX_{k}],\\
				\bar{\bY}_k &= \mathbf{H}(\hat{\bX}_{k})-\by_{k}.
			\end{aligned}
		\end{equation*}
		Then the optimization objective \eqref{eq:penalty MMD-1} has the  global minimum, if
		\begin{equation}\label{eq:lineartrasport}
			\mathbf{T} = \mathbb{E}\left[\bar{\bX}_k\bar{\bY}_k^{\top}\right]\left\{\mathbb{E}\left[\bar{\bY}_k\bar{\bY}_k^{\top}\right]+\mathbb{E}[\boldsymbol{\varepsilon}_{k}\boldsymbol{\varepsilon}_{k}^{\top}]\right\}^{-1}.
		\end{equation}
	\end{thm}
	
	\begin{proof}
		By plugging  \eqref{eq:linear transport form} into \eqref{eq:penalty MMD-1}, the optimization objective can be rewriten  as
		\begin{equation*}
			\begin{aligned}
				\operatorname{MMD}[p, q;\mathcal{F}] = & \mathbb{E}[(\by_{k} + \boldsymbol{\varepsilon}_{k} - \mathbf{H}(\hat{\bX}_{k}))^{\top}\mathbf{T}^{\top}\mathbf{T}(\by_{k} + \boldsymbol{\varepsilon}_{k} - \mathbf{H}(\hat{\bX}_{k}))]\\
				&+2\mathbb{E}[\hat{\bX}_{k}^{\top}\mathbf{T}(\by_{k} + \boldsymbol{\varepsilon}_{k} - \mathbf{H}(\hat{\bX}_{k}))]\\
				&-2\mathbb{E}[\bX_k^{\top}\mathbf{T}(\by_{k} + \boldsymbol{\varepsilon}_{k} - \mathbf{H}(\hat{\bX}_{k}))]\\
				&+ C.
			\end{aligned}
		\end{equation*}
		We take $\frac{\partial \operatorname{MMD}[p, q;\mathcal{F}]}{\partial \mathbf{T}} = 0$. According to the property of matrix derivatives
		\begin{equation*}
			\frac{\partial \mathbf{a}^T \bx \mathbf{b}}{\partial \bx}=\mathbf{a b}^T, \qquad \frac{\partial \mathbf{b}^T \bx^T \bx \mathbf{c}}{\partial \bx}=\bx\left(\mathbf{b} \mathbf{c}^T+\mathbf{c b}^T\right),
		\end{equation*}
		we have 
		\begin{equation*}
			\begin{aligned}
				\mathbf{T}\mathbb{E}[(\by_{k} + \boldsymbol{\varepsilon}_{k} - \mathbf{H}(\hat{\bX}_{k}))(\by_{k} + \boldsymbol{\varepsilon}_{k} - \mathbf{H}(\hat{\bX}_{k}))^{\top}] =& \mathbb{E}[\hat{\bX}_{k}(\by_{k} + \boldsymbol{\varepsilon}_{k} - \mathbf{H}(\hat{\bX}_{k}))^{\top}]\\
				&-\mathbb{E}[{\bX}_{k}(\by_{k} + \boldsymbol{\varepsilon}_{k} - \mathbf{H}(\hat{\bX}_{k}))^{\top}].
			\end{aligned}
		\end{equation*}
		Because  $\pi^{\mathcal{T}_k}_{\bX_{k}\mid \bY_{1:k}}$ and $\pi_{\bX_{k}\mid \bY_{1:k}}$ in \eqref{eq:loss} are two different distributions,  the corresponding random variables $\bX_{k}^{\mathcal{T}_k}$ and $\bX_k$ are set to be independent. Thus
		\begin{equation*}
			\begin{aligned}
				\mathbf{T}\left\{\mathbb{E}[(\by_{k} - \mathbf{H}(\hat{\bX}_{k}))(\by_{k} - \mathbf{H}(\hat{\bX}_{k}))^{\top}] + \mathbb{E}\left[\boldsymbol{\varepsilon}_{k}\boldsymbol{\varepsilon}_{k}^{\top}\right]\right\} =& \mathbb{E}[\hat{\bX}_{k}(\by_{k} - \mathbf{H}(\hat{\bX}_{k}))^{\top}]\\
				&-\mathbb{E}[{\bX}_{k}]\mathbb{E}[\by_{k} - \mathbf{H}(\hat{\bX}_{k})]^{\top}.
			\end{aligned}
		\end{equation*}
		Let $\bar{\bX}_k := \hat{\bX}_{k}-\mathbb{E}[\bX_{k}], \bar{\bY}_k = \mathbf{H}(\hat{\bX}_{k})-\by_{k}$. Then we have 
		\begin{equation*}
			\mathbf{T} = \mathbb{E}\left[\bar{\bX}_k\bar{\bY}_k^{\top}\right]\left\{\mathbb{E}\left[\bar{\bY}_k\bar{\bY}_k^{\top}\right]+\mathbb{E}[\boldsymbol{\varepsilon}_{k}\boldsymbol{\varepsilon}_{k}^{\top}]\right\}^{-1}.
		\end{equation*}
		Proof is completed.
	\end{proof}
	
	{Since true states $\bx_k$ and $\by_k=\mathbf{H}_k(\bx_k)$ are not avaliable, we replace $\by_k$ by $\by_k^o$.}  By  \eqref{eq:particle filter posterior},  the expectation  of posterior $\mathbb{E}\left[\bX_k\right] \approx \sum_{i=1}^N w_i \hat{\bx}_{k}^{(i)}$. The  ensemble approximation of \eqref{eq:lineartrasport} are computed  by
	\begin{align}
		\mathbb{E}\left[\bar{\bX}_k\bar{\bY}_k^{\top}\right] &\approx \frac{1}{N-1}\sum_{i=1}^N\left(\hat{\bx}_{k}^{(i)}-\mathbb{E}\left[\bX_k\right]\right)\left(\mathbf{H}(\hat{\bx}_{k}^{(i)})-\by^o_{k}\right)^{\top},\label{eq:empirical T-1}\\
		\mathbb{E}\left[\bar{\bY}_k\bar{\bY}_k^{\top}\right] &\approx \frac{1}{N-1}\sum_{i=1}^N\left(\mathbf{H}(\hat{\bx}_{k}^{(i)})-\by^o_{k}\right)\left(\mathbf{H}(\hat{\bx}_{k}^{(i)})-\by^o_{k}\right)^{\top},\\
		\mathbb{E}[\boldsymbol{\varepsilon}_{k}\boldsymbol{\varepsilon}_{k}^{\top}] &\approx\frac{1}{N-1}\sum_{i=1}^N\left(\boldsymbol{\varepsilon}^{(i)}_{k}\boldsymbol{\varepsilon}^{(i)}_{k}\right)^{\top},\label{eq:empirical T-3}
	\end{align}
	where $\{\boldsymbol{\varepsilon}^{(i)}_{k}\}_{i=1}^N$ are drawn from observation noise.
	
	
	\begin{rem}
		Equation \eqref{eq:lineartrasport} shows that the ensemble transport filter with a variance penalty MMD loss has a similar form compared to the  EnKF with a linear assumption and linear kernel. Denote
		\begin{equation*}
			\tilde{\bX}_k = \hat{\bX}_{k+1}-\mathbb{E}[\hat{\bX}_{k+1}],\qquad
			\tilde{\bY}_k = \mathbf{H}(\hat{\bX}_{k+1})-\mathbb{E}\left[\mathbf{H}(\hat{\bX}_{k+1})\right].
		\end{equation*}
		the Kalman gain of EnKF can be rewritten as
		\begin{equation}\label{eq:kalman gain}
			\mathbf{K} = {\mathbb{E}\left[\tilde{\bX}_k\tilde{\bY}_k^{\top}\right]}\left\{\mathbb{E}\left[\tilde{\bY}_k\tilde{\bY}_k^{\top}\right]+\mathbb{E}[\boldsymbol{\varepsilon}_{k+1}\boldsymbol{\varepsilon}_{k+1}^{\top}]\right\}^{-1}.
		\end{equation}
		The main difference between EnTranF and EnKF  is that  the means of $\hat{\bX}_{k+1}$ and $\mathbf{H}(\hat{\bX}_{k+1})$  are approximated by  $\mathbb{E}[{\bX}_{k+1}]$ and $\by_k$, respectively, in EnTranF.   This  drives  the prior particles towards the posterior mean as dictated by transport map \eqref{eq:linear transport form}.
	\end{rem}
	
	\begin{rem}
		When the RKHS kernel takes $k(\bx, \bx')=\bx^{\top}\bx'+1$ and take regularization parameter to be $\lambda=0$,   we follow the proof of  \ref{thm:EnTranFp LL} and get 
		\begin{equation}\label{eq:mmd slope}
			\mathbf{T} = \mathbb{E}[\bar{\bX}_k]\mathbb{E}\left[\bar{\bY}_k\right]^{\top}\left\{\mathbb{E}\left[\bar{\bY}_k\right]\mathbb{E}\left[\bar{\bY}_k\right]^{\top}\right\}^{-1}.
		\end{equation}
		Formally, \eqref{eq:lineartrasport} provides a complete slope in the statistical least squares sense, while \eqref{eq:mmd slope} gives the geometric slope from the prior mean to the posterior mean. However, when the number of particles is small, \eqref{eq:mmd slope} may be numerically unstable in computing the matrix inversion.
	\end{rem}
	
	\section{Numerical Results}\label{sec:4}
	In this section, we will use a  few numerical examples  to illustrate  the advantage of the EnTranF over  ensemble Kalman filtering in the  data assimilation.  A  static inverse problem is  implemented to show the capability of EnTranF in approximating posterior distributions in \ref{sec:4.1}. 
	{In \ref{sec:4.2}, we discuss the performance of EnTranF in nonlinear system by tracking Double-Well system. For strongly nonlinear cases, we consider the filtering problem of the Lorenz63 system with partial observations in \ref{sec:4.3}. Finally, the performance of EnTranF for tracking high-dimensional systems is demonstrated in \ref{sec:4.4}.}
	
	We use component average Root-Mean-Squared Error (RMSE) to evaluate the performance of the assimilation method, which is defined by 
	\[
	\text{RMSE}_k := \|\bar{\bx}_k - \bx_k^*\|_2 / \sqrt{n}
	\]
	at any assimilation step $k$. Here $\bar{\bx}_k\in \mathbb{R}^n$ is the (ensemble) mean state generated by the data assimilation method, and $\bx_k^*\in \mathbb{R}^n$ is the reference state. Considering all assimilation windows throughout the filtering process, we compute  the time-averaged RMSE. Meanwhile, to eliminate the influence of randomness, we define the final performance metric RMSE as its expectation, i.e.,
	\begin{equation}\label{eq:RMSE}
		\text{RMSE} = \mathbb{E}\left[\frac{1}{T}\sum\limits_{k=1}^{T}\|\bar{\bx}_k - \bx_k^*\|_2 / \sqrt{n}\right].
	\end{equation}
	Moreover,  Ensemble Spread(ENS) is used to measure the concentration of the ensemble samples, which is computed by $\text{ENS}_k = \left[\mathop{\text{tr}}(\mathbf{C}_k^a)/n\right]^{1/2}$, where $\mathbf{C}_k^a$ is the analysis ensemble covariance matrix at time step $k$. Similarly, a time average is used to monitor the ENS over  the whole assimilation process.
	\begin{equation}
		\text{ENS} = \mathbb{E}\left[\frac{1}{T}\sum\limits_{k=1}^{T}\left[\mathop{\text{tr}}(\mathbf{C}_k^a)/n\right]^{1/2}\right].
	\end{equation}
	We will use  Coverage Probability (CP) to fully  assess RMSE and Ensemble Spread. CP refers to the probability, with which  a confidence interval of the filtered state encompasses the reference. A higher CP indicates better consistency between the filtered and reference states. Here, we define CP by
	\begin{equation}\label{eq:CP}
		\text{CP} = \mathbb{E}\left[\frac{1}{n}\sum_{i=1}^n\left(\frac{1}{T}\sum_{k=1}^T \mathbb{I}_{\left|\bar{\bx}_{k, i}-\bx^*_{k,i}\right|/\sqrt{\mathbf{C}_{k, (i,i)}^a}\le t_{\alpha}}\right)\right],
	\end{equation}
	where $\mathbb{I}$ is indicator function and $t_{\alpha}$ is $\alpha$ quantiles of standard normal distribution. We take $\alpha=0.025$ and compute the expectation in \eqref{eq:RMSE}-\eqref{eq:CP} using 20 repeated tests.
	
	\subsection{Static inverse problem}\label{sec:4.1}
	In this subsection, we consider two static inverse problems to examine the validity of the transport map from the prior to the posterior distribution based on the MMD loss function. We will compare the EnTranF with the EnKF and examine how well it approximates the true posterior distribution. 
	
	{Start with a one-dimensional example, and suppose $X$ is a random variable with Gaussian prior $\mathcal{N}(0.5, 1)$ and Gaussian observation noise $\epsilon\sim \mathcal{N}(0, 0.5^2)$. The nonlinear observation operator is considered as
		${H}(x) = 2x^3+x$.
		The task is to estimate the Bayesian posterior of the $X$ given values of observations $y^{o}=1.2$. }
	In EnTranF and EnKF, particles are generated to approximate the posterior distribution.
	Then, a kernel density estimate method is used to evaluate the density function of ensembles. For reference, based on Bayes' rule
	$$
	\pi_{posterior}(x\mid y) \propto \pi_{prior}(x)\times\pi_{likelihood}(y\mid x),
	$$
	The exact posterior density can be calculated using the numerical integral method. 
	
	\begin{figure}[!htbp]
		\centering
		\subfigure[\bfseries{Linear and $\text{bw}=10^{-3}$}]{
			\begin{minipage}[t]{0.3\textwidth}
				\centering
				\includegraphics[width=\textwidth]{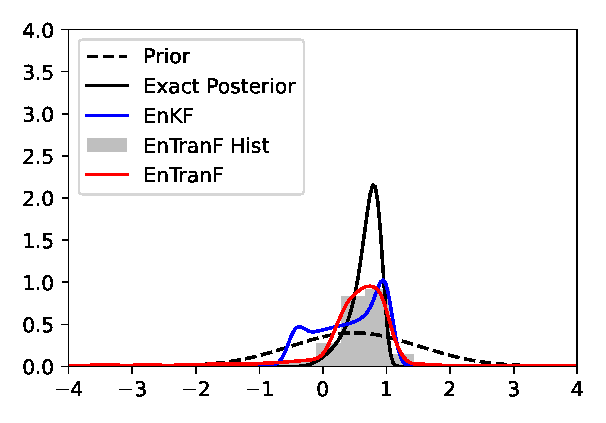}
			\end{minipage}
		}
		\subfigure[Linear and $\text{bw}=10^{0}$]{
			\begin{minipage}[t]{0.3\textwidth}
				\centering
				\includegraphics[width=\textwidth]{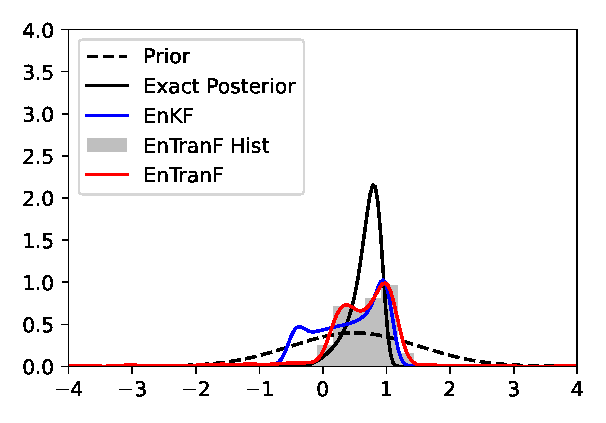}
			\end{minipage}
		}
		\subfigure[Linear and $\text{bw}=10^{3}$]{
			\begin{minipage}[t]{0.3\textwidth}
				\centering
				\includegraphics[width=\textwidth]{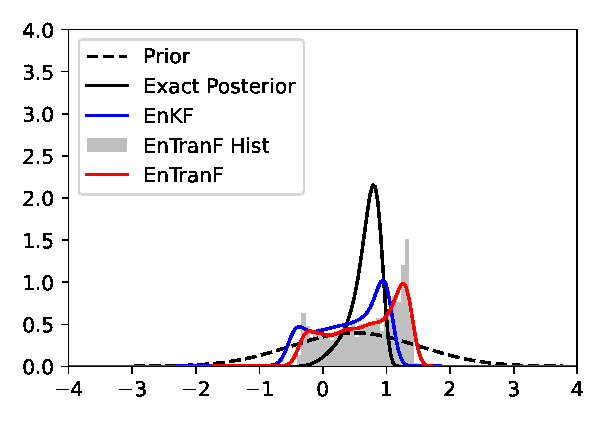}
			\end{minipage}
		}\\
		\subfigure[\bfseries{Nonlinear and $\text{bw}=10^{-3}$}]{
			\begin{minipage}[t]{0.3\textwidth}
				\centering
				\includegraphics[width=\textwidth]{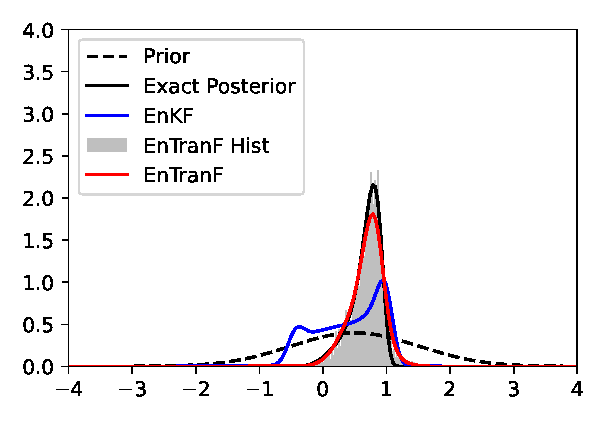}
			\end{minipage}
		}
		\subfigure[{Nonlinear and $\text{bw}=10^{0}$}]{
			\begin{minipage}[t]{0.3\textwidth}
				\centering
				\includegraphics[width=\textwidth]{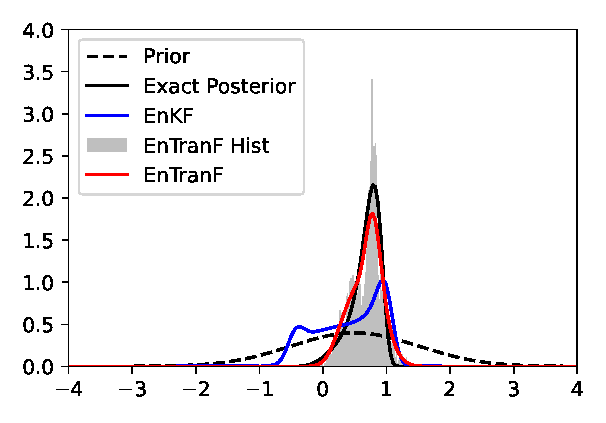}
			\end{minipage}
		}
		\subfigure[Nonlinear and $\text{bw}=10^{3}$]{
			\begin{minipage}[t]{0.3\textwidth}
				\centering
				\includegraphics[width=\textwidth]{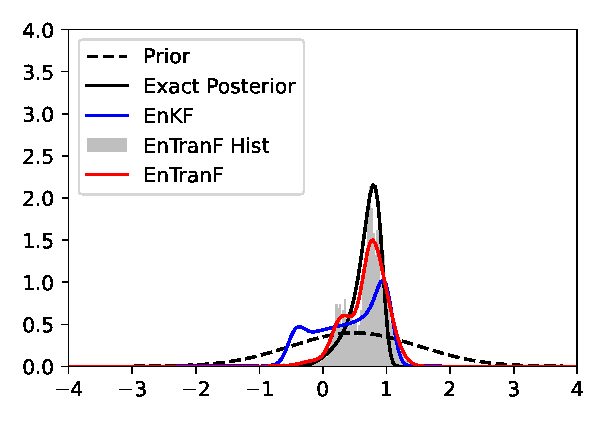}
			\end{minipage}
		}
		\caption{Posterior approximation of EnTranF with Linear transport map (Panel (a)-(c)) and Nonlinear transport map (Panel (d)-(f))  when taking different bandwidths ($10^{-3}$, $10^{0}$ and $10^{3}$).}
		\label{fig:1d-static}
	\end{figure}
	
	We  select  Gaussian kernel for MMD loss with three different bandwidth values $10^{-3}$, $10^{0}$, and $10^{3}$. Linear and Nonlinear transport maps are used  to approximate posterior in EnTranF, and the  results are shown in Figure \ref{fig:1d-static}. Larger bandwidth $\text{bw}=10^{3}$ leads to more accurate  estimates of the mean compared to EnKF, but it fails to capture the morphology of the posterior distribution. When a smaller bandwidth ($\text{bw}=10^{0}$ and $\text{bw}=10^{-3}$)  is given, estimation of the posterior distribution is more concentrated in the high-probability region of the true posterior distribution. EnTranF becomes more sensitive to the local details of the posterior distribution when a smaller bandwidth is chosen.
	
	The nonlinear observation operator leads to  the non-Gaussian of the posterior distribution. Both EnKF and EnTranF with linear transport maps fail to approximate the posterior distribution. As shown in Figure \ref{fig:1d-static}, EnTranF achieves a significant improvement in approximating posterior distribution with a nonlinear transport map. This could be caused by the substantially different sensitivity of observations to the model state for each ensemble particle in the  nonlinear assimilation problem. A linear form of transport map may not well  approximate  a non-Gaussian posterior.
	
	At the same time, we will examine the performance of EnTranF, which  optimizes the MMD with a variance penalty in approximating the posterior distribution.   The penalted EnTranF is  compared to the case without a penalty term on a two-dimensional state space $\bx = [\bx_1, \bx_2]$.  {We assume Gaussian prior $\mathcal{N}\left(\left[\begin{matrix}
			0.5\\
			0.5
		\end{matrix}\right], \left[\begin{matrix}
			1& 0\\
			0& 1
		\end{matrix}\right]\right)$ and the nonlinear observation $y^o=0.8$ with operator  $\mathbf{H}(\bx) = \bx_1^3 + \bx_2$ and additive Gaussian noise $\epsilon\sim \mathcal{N}(0, 0.5^2)$.}
	\begin{figure}[!ht]
		\centering
		\subfigure[Exact Posterior]{
			\begin{minipage}[t]{0.22 \textwidth}
				\centering
				\includegraphics[width=\textwidth]{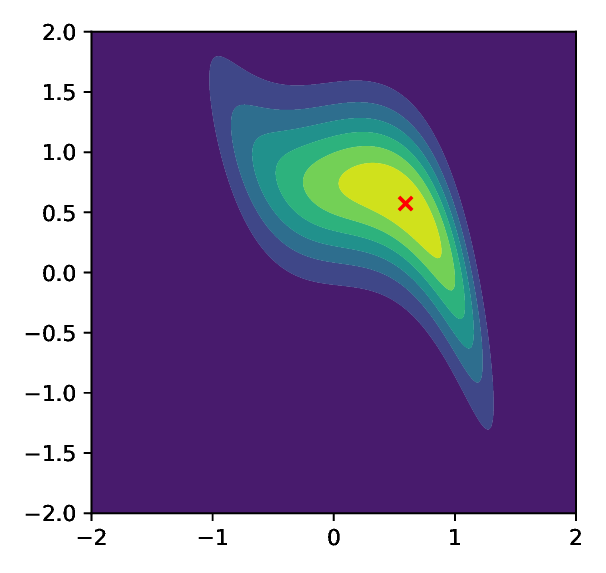}
			\end{minipage}
		}
		\subfigure[EnKF]{
			\begin{minipage}[t]{0.22\textwidth}
				\centering
				\includegraphics[width=\textwidth]{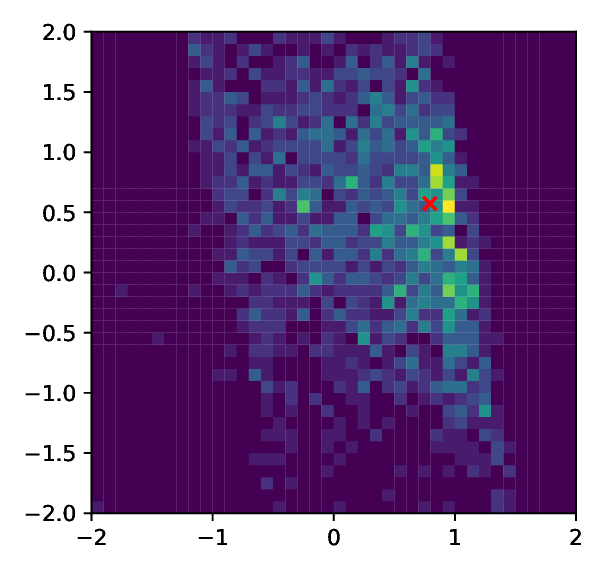}
			\end{minipage}
		}
		\subfigure[EnTranF]{
			\begin{minipage}[t]{0.22\textwidth}
				\centering
				\includegraphics[width=\textwidth]{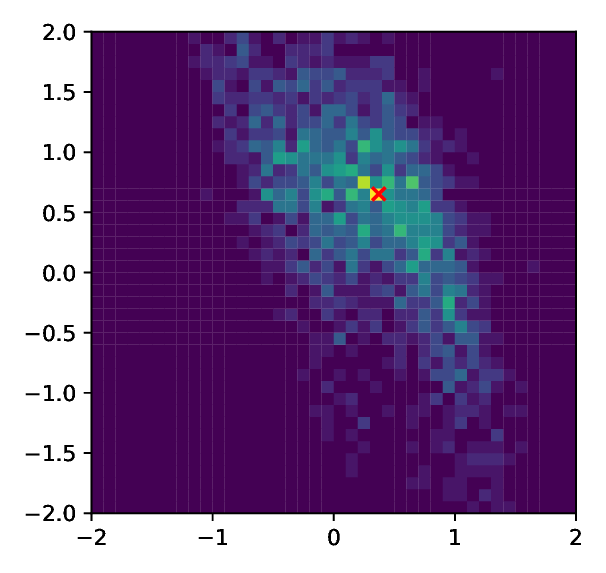}
			\end{minipage}
		}
		\subfigure[EnTranFp]{
			\begin{minipage}[t]{0.22\textwidth}
				\centering
				\includegraphics[width=\textwidth]{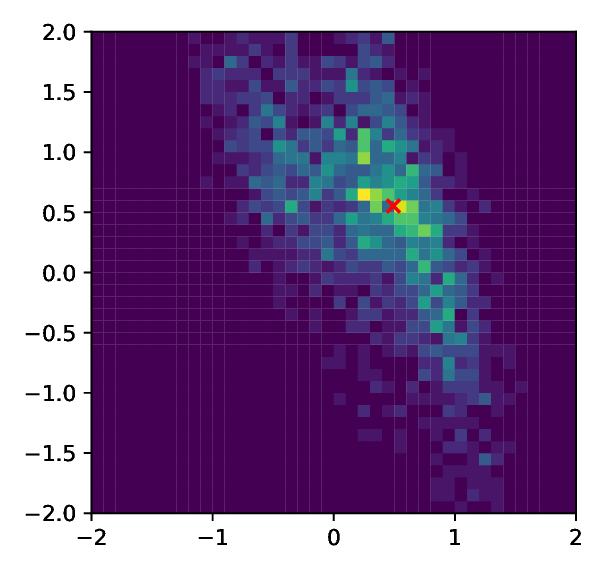}
			\end{minipage}
		}
		\caption{2-d Histogram of ensemble particles generated by EnKF, EnTranF and EnTranFp. Red "$\times$" represents the MAP estimation of the corresponding method.}
		\label{fig:2d-static}
	\end{figure}
	
\begin{figure}[!ht]
		\centering
		\includegraphics[width=.5\textwidth]{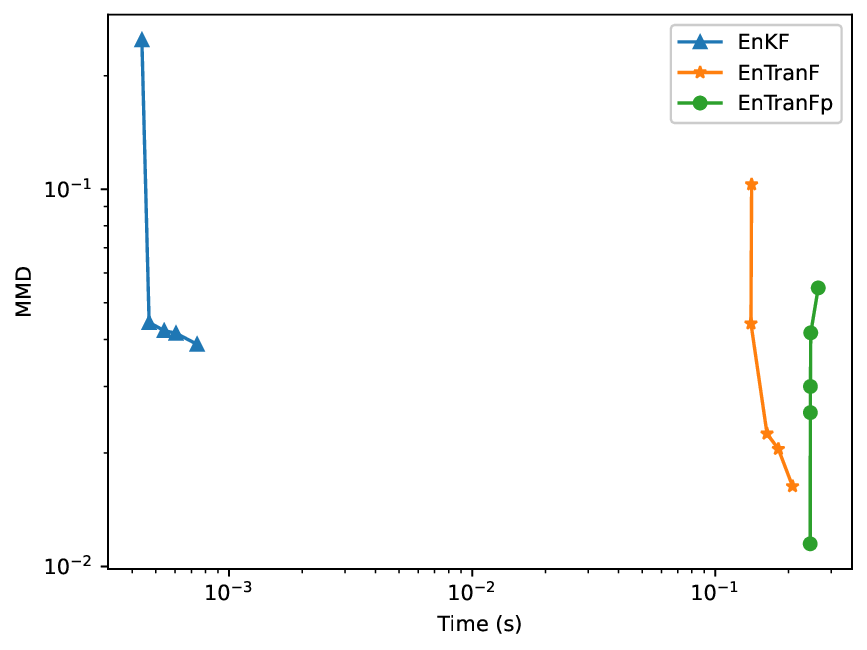}
		\caption{log-log plot of computation time and MMD of EnKF, EnTranF and EnTranFp. }
		\label{fig:time}
	\end{figure}

	From Figure \ref{fig:2d-static}, EnTranF and EnTranFp achieve accurate  estimate of the posterior distribution, while EnKF has a  significant bias. The particles of EnTranFp are more concentrated than EnTranF, and the variance penalty effectively improves the approximation of the posterior distribution in high-probability regions. {Table \ref{tab:estimation post RMSE} lists the average RMSE and ENS of different filtering methods with different ensemble sizes.} The EnTranFp achieves the lowest RMSE and exhibits the highest stability across both ensemble sizes (200 and 400). EnTranF shows slightly improved performance over EnTranFp at larger ensemble sizes (800), both variants significantly outperform EnKF in terms of RMSE. {As for the ENS, we use a particle filter with 10,000 particles as a reference benchmark. As shown, the ENS of EnTranF and EnTranFp are notably closer to this reference than those of the EnKF.} These results highlight the effectiveness of our approach in generating more accurate posterior estimates. Furthermore, the advantage of EnTranFp shows the importance of the variance penalty in enhancing both the accuracy and stability of the transport map. {Figure \ref{fig:time} shows the relationship between the distribution approximation accuracy and computational time for EnKF, EnTranF, and EnTranFp. EnKF requires less computational time but has lower distribution approximation accuracy; EnTranF, due to the need to optimize the loss function, consumes more computational time, but achieves higher distribution approximation accuracy.}
	
	\begin{table}[!ht]
		\caption{ Average RMSE and ENS of EnKF, EntranF and EnTranFp for the 2d static inverse problem with different ensemble size. }
		\begin{center}
			{Results of RMSE}\\
			\vspace{5pt}
			\begin{tabular}{ccccccc}
				\hline
				&EnKF & EnTranF & EnTranFp\\
				\hline
				{200}& 0.1363($\pm$0.0862) & 0.1377($\pm$0.0732) & {\bfseries 0.1255($\pm$0.0592)}\\
				\hline
				{400}& 0.1543($\pm$0.0510) & 0.0962($\pm$0.0554) & {\bfseries 0.0878($\pm$0.0498)}\\
				\hline
				{800}& 0.1329($\pm$0.0381) & {\bfseries 0.0702($\pm$0.0331)} & 0.0742($\pm$0.0350)\\
				\hline
			\end{tabular}\\
			\vspace{1em}
			{{Results of ENS}\\
			\vspace{5pt}
			\begin{tabular}{ccccccc}
				\hline
				&Ref.&EnKF & EnTranF & EnTranFp\\
				\hline
				{200} & 0.6076
				 & 0.8340($\pm$0.0387) & 0.8197($\pm$0.0728) & {\bfseries 0.7616($\pm$0.0441)}\\
				\hline
				{400} & 0.6076 & 0.8216($\pm$0.0225) & 0.8059($\pm$0.0664) & {\bfseries 0.7466($\pm$0.0455)}\\
				\hline
				{800} & 0.6076 & 0.8332($\pm$0.0252) & {0.7636($\pm$0.0286)} & {\bfseries 0.7306($\pm$0.0221)}\\
				\hline
			\end{tabular}}
		\end{center}
		\label{tab:estimation post RMSE}
	\end{table}
	
	\subsection{Double-Well potential}\label{sec:4.2}
	Double-well potential is a common form of potential energy in physics and chemistry. It is typically used to describe the movement of molecules or particles in bistable systems, often characterized by two local energy minima that allow particles to oscillate between these two points under random perturbations. This potential form is essential in describing various physical and chemical phenomena, such as conformational transitions within molecules and spin flips in magnetic materials. This subsection mainly demonstrates the performance of the EnTranF method for nonlinear filtering problem in tracking the Double-Well potential system. The dynamical model and observation operator are given by
	\begin{align}
		\mathrm{d} x &= (x - x^3)\mathrm{d} t + \gamma{\mathrm{d} W_t},\label{eq:evo1dDoubleWell}\\
		z &= 0.1x^2 + \sin(x) + \epsilon, \label{eq:ob1dDoubleWell}
	\end{align}
	where $W_t$ is a standard brownian motion and $\epsilon$ is Gaussian noise with $\epsilon \sim \mathcal{N}(0, \sigma^2)$.  We take $\gamma = 0.8$ and $\sigma=0.5$. In this example, the fourth-order Runge-Kutta method is used to simulate the ODE system \eqref{eq:evo1dDoubleWell} with a constant discrete step size $\Delta t = 0.01$. Results of EnTranF with linear and nonlinear transport map are demonstrated. { For nonlinear transport map, a shallow DNN comprising $L=3$ layers with width $\{1, 10, 1\}$ and tanh activation functions is used to define the nudging term of the nonlinear transport map. The bandwidth of MMD was set to be the median of all $L_2$ distance between each pair of particles.}
	
	For simplicity of notation, we use the following abbreviation to denote the ensemble transport filter:
	\begin{itemize}
		\item EnTranF(L-L): the ensemble transport filter with linear transport map and linear kernel function;
		\item EnTranF(L-G): the ensemble transport filter with linear transport map and Gaussian kernel function;
		\item EnTranF(N-L): the ensemble transport filter with nonlinear transport map and linear kernel function;
		\item EnTranF(N-G): the ensemble transport filter with nonlinear transport map and Gaussian kernel function;
	\end{itemize}
	
\begin{figure}[!htbp]
		\centering
		\subfigure{
			\begin{minipage}[t]{0.46\textwidth}
				\centering
				\includegraphics[width=\textwidth]{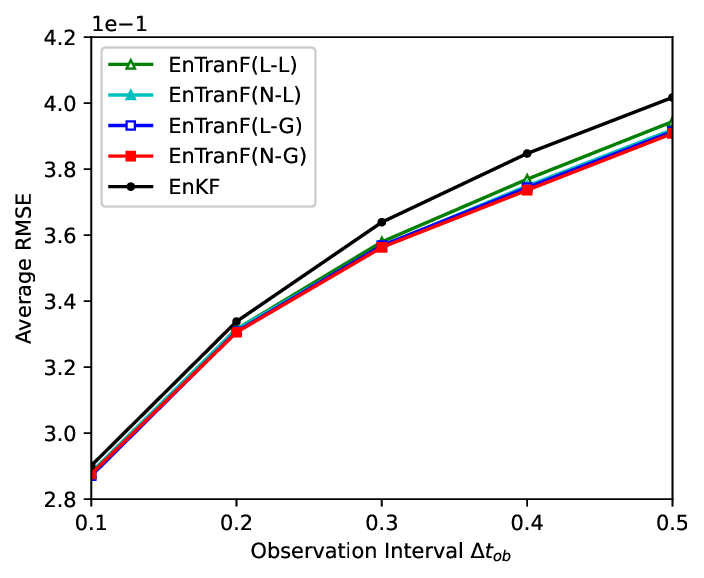}
			\end{minipage}
		}
		\subfigure{
			\begin{minipage}[t]{0.46\textwidth}
				\centering
				\includegraphics[width=\textwidth]{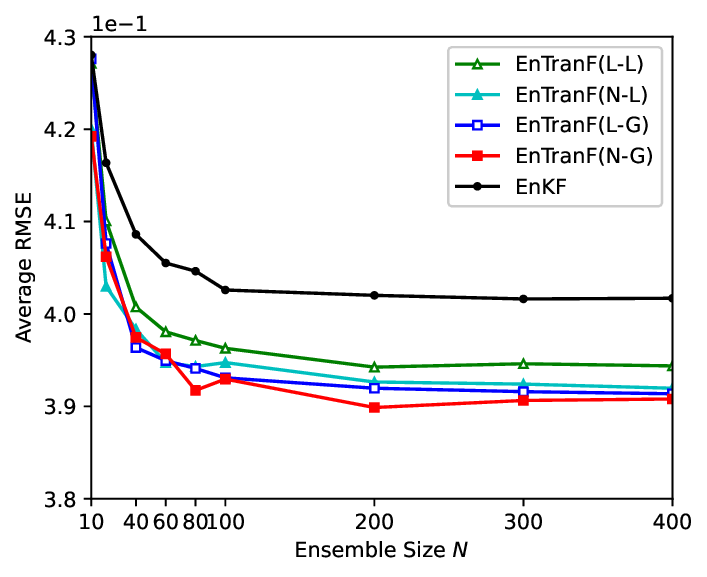}
			\end{minipage}
		}
		\caption{Average RMSE of Double-Well system for ensemble size $N=400$ as a function of observation interval (Left), and for observation interval $\Delta t_{ob}=0.5$ as a function of ensemble size (Right).}
		\label{fig:RMSE DoubleWell}
	\end{figure}

	Figure \ref{fig:RMSE DoubleWell} compares the average RMSE between EnTranF and EnKF with different observation intervals and ensemble sizes. 
	
	The left panel of Figure \ref{fig:RMSE DoubleWell} indicates that EnTranF exhibits significant improvement on average RMSE compared to EnKF at larger observation intervals $\Delta_{ob}>0.2$ while performing similarly at smaller observation intervals. 
	The nonlinearity of the filtering problem increases as the observation interval increases, leading to a larger difference in RMSE between EnKF and EnTranF. 
	EnTranF(N-G)  demonstrates the best performance. However, compared to EnTranF(N-L) and EnTranF(L-G), the difference on average RMSE is small. 
	{The right panel of Figure \ref{fig:RMSE DoubleWell} shows that EnTranF achieves a smaller average RMSE compared to EnKF as the increase of ensemble size}. 
	Nonlinear transport maps exhibit better performance on average RMSE than linear transport maps. 
	However, the difference is insignificant when taking the Gaussian kernel function. 
	The EnTranF based on Gaussian kernel function keeps sufficient posterior information in filtering. 
	
\begin{figure}[!htbp]
		\centering
		\subfigure{
			\begin{minipage}[t]{0.45\textwidth}
				\centering
				\includegraphics[width=\textwidth]{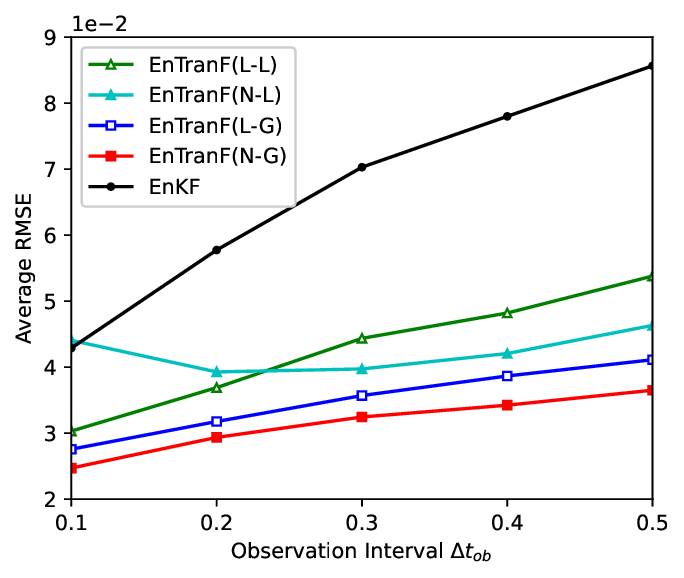}
			\end{minipage}
		}
		\subfigure{
			\begin{minipage}[t]{0.45\textwidth}
				\centering
				\includegraphics[width=\textwidth]{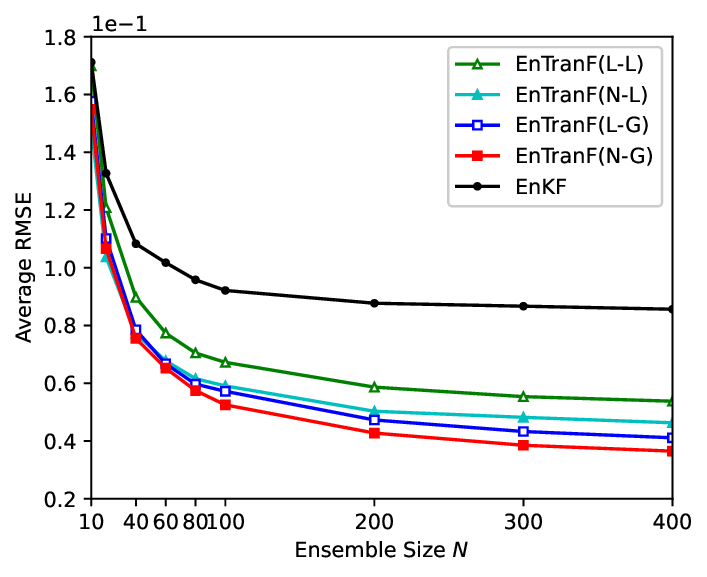}
			\end{minipage}
		}
		\caption{Take posterior mean of PF with 10,000 particles as reference, Average RMSE of Double-Well system for ensemble size $N=400$ as a function of observation interval (Left), and for observation interval $\Delta t_{ob}=0.5$ as a function of ensemble size  (Right).}
		\label{fig:RMSE DoubleWell Post}
	\end{figure}

	Since the DoubleWell potential represents a low-dimensional system, employing a particle filter with 10,000 particles as a benchmark posterior allows us to illustrate the approximate performance of EnTranF on the true posterior filtering distribution. To this end, we assess the tracking performance of the EnTranF using metrics identical to those employed to evaluate the estimation of the true state. Results are shown  in Figure \ref{fig:RMSE DoubleWell Post}.
	
	When measuring the performance of EnTranF in tracking the posterior mean, it exhibits a significant improvement on average RMSE compared to EnKF. From the left panel of Figure \ref{fig:RMSE DoubleWell Post}, it can be observed that EnTranF shows an average improvement of $38.16\%$ compared to EnKF across various observation intervals. At the same time, as the observation interval increases, the rate of increase in RMSE for EnTranF is lower than that for EnKF. This implies  that EnTranF has better robustness as the nonlinearity increases. EnTranF(N-L) exhibits poor performance when the observation interval is 0.1, which may stem from inadequate optimization of the nonlinear transport map and getting trapped in local minima. As shown in the right panel of Figure \ref{fig:RMSE DoubleWell Post}, EnTranF gradually approaches reference solution as the number of particles increases. However, the average RMSE of EnKF do not decrease when the ensemble size becomes larger than $200$. 
	
	\begin{figure}[!htbp]
		\centering
		\subfigure{
			\begin{minipage}[t]{0.45\textwidth}
				\centering
				\includegraphics[width=\textwidth]{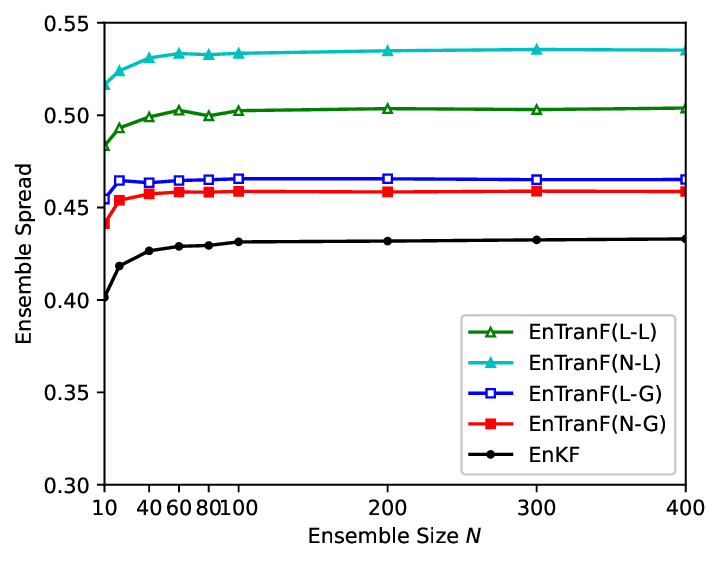}
			\end{minipage}
		}
		\subfigure{
			\begin{minipage}[t]{0.45\textwidth}
				\centering
				\includegraphics[width=\textwidth]{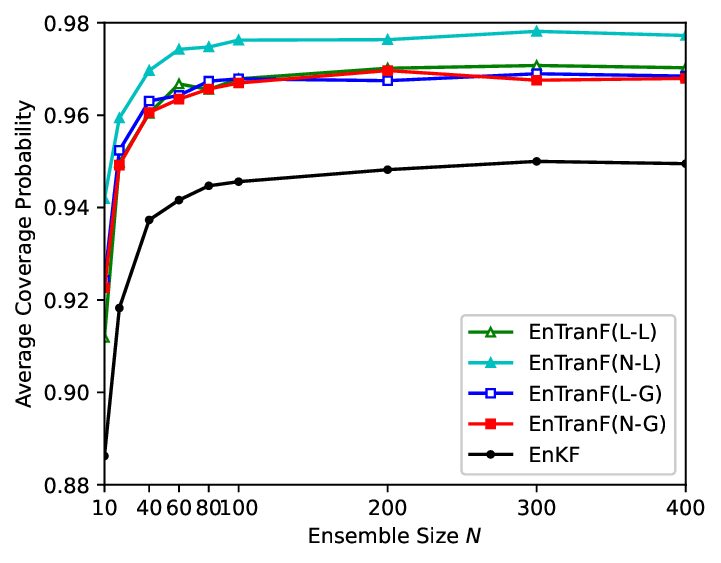}
			\end{minipage}
		}
		\caption{Left Panel: Spread of Double-Well system for observation interval $\Delta t_{ob}=0.5$ as a function of ensemble size. Right panel: Average coverage probability of Double-Well system for observation interval $\Delta t_{ob}=0.5$ as a function of ensemble size}
		\label{fig:Spread DoubleWell}
	\end{figure}
	
\begin{figure}[!htbp]
		\centering
		\subfigure[]{
			\begin{minipage}[t]{0.45\textwidth}
				\centering
				\includegraphics[width=\textwidth]{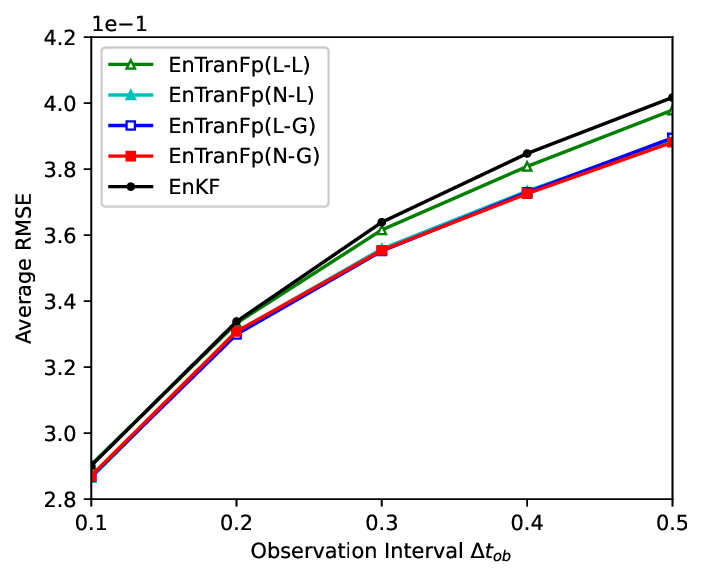}
			\end{minipage}
		}
		\subfigure[]{
			\begin{minipage}[t]{0.45\textwidth}
				\centering
				\includegraphics[width=\textwidth]{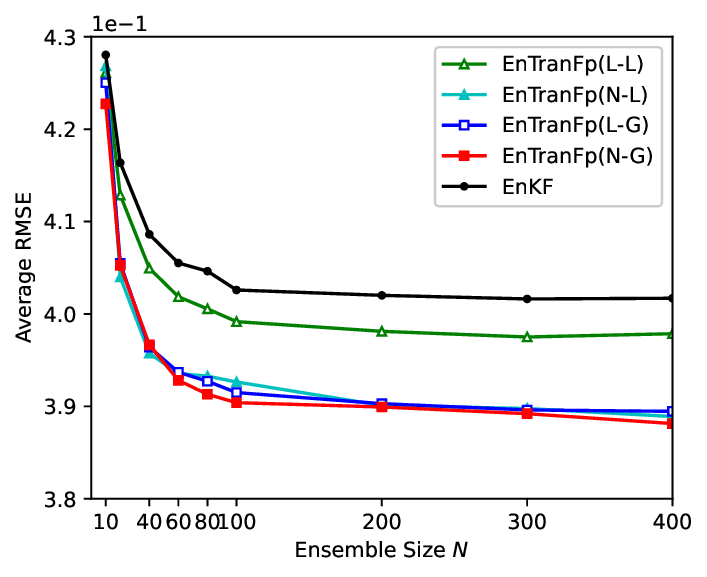}
			\end{minipage}
		}\\
		\subfigure[]{
			\begin{minipage}[t]{0.45\textwidth}
				\centering
				\includegraphics[width=\textwidth]{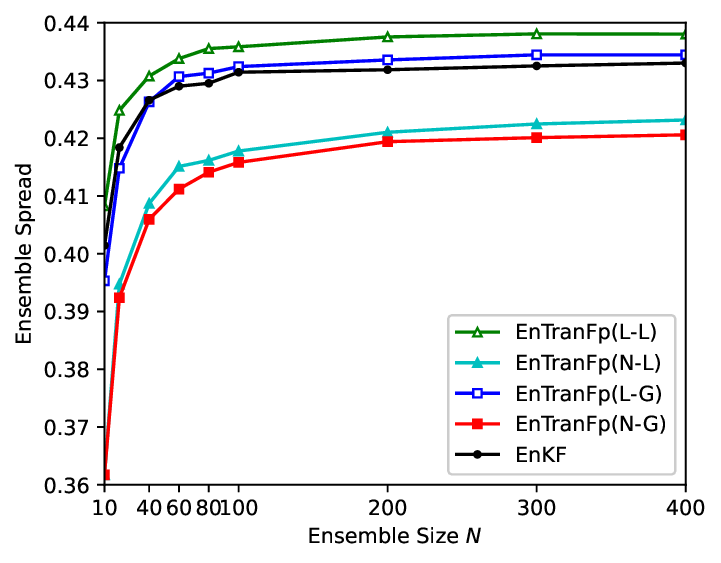}
			\end{minipage}
		}
		\subfigure[]{
			\begin{minipage}[t]{0.45\textwidth}
				\centering
				\includegraphics[width=\textwidth]{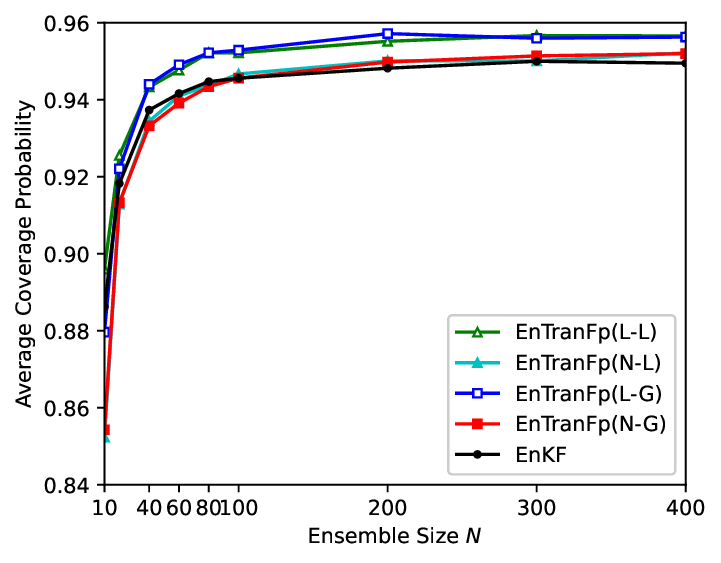}
			\end{minipage}
		}
		\caption{EnTranF with variance penalty: Average RMSE of Double-Well system for ensemble size $N=400$ as a function of observation interval (Panel (a)), and for observation interval $\Delta t_{ob}=0.5$ as a function of ensemble size (Panel (b)). Spread of Double-Well system for observation interval $\Delta t_{ob}=0.5$ as a function of ensemble size (Panel (c)). Average coverage probability of Double-Well system for observation interval $\Delta t_{ob}=0.5$ as a function of ensemble size  (Panel (d))} 
		\label{fig:RMSE Spread DoubleWell Penalty}
	\end{figure}

	While EnTranF demonstrates excellent performance in state estimation, its performance on ensemble spread is poor, leading to inadequate particle dispersion in ensemble filtering methods  (see Figure \ref{fig:Spread DoubleWell}). EnTranF also suffers from numerical overflow when the number of particles is low. After incorporating a variance penalty term, EnTranF can effectively address these issues. Figure \ref{fig:RMSE Spread DoubleWell Penalty} shows  the RMSE and Spread of EnTranFp for different observation intervals and ensemble sizes.
	
\begin{figure}[!ht]
		\centering
		\includegraphics[width=.45\textwidth]{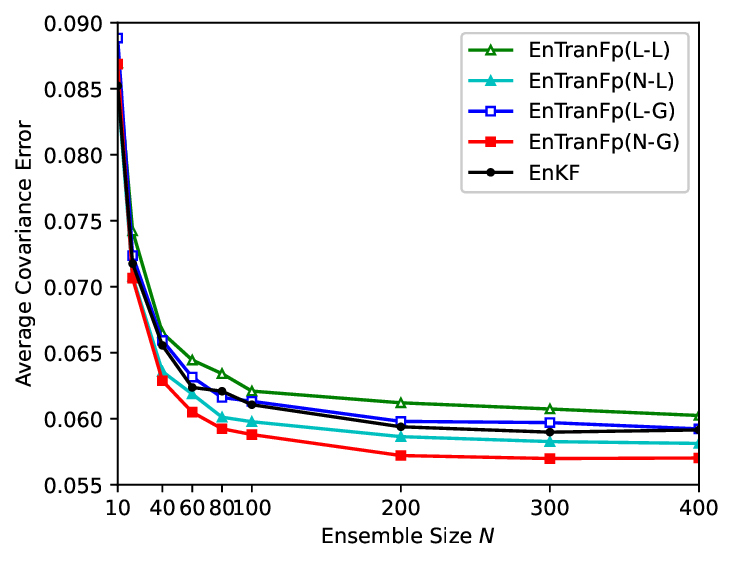}
		\caption{EnTranF with variance penalty: Average covariance error of Double-Well system for observation interval $\Delta t_{ob}=0.5$ as a function of ensemble size.}
		\label{fig:cov_error_DW}
	\end{figure}

	From Figure \ref{fig:RMSE DoubleWell} and Figure \ref{fig:RMSE Spread DoubleWell Penalty} (a)-(b), the average RMSE of EnTranFp remains consistent across different observation intervals and ensemble sizes. The EnTranFp maintains the accurancy of EnTranF in state estimation.
	EnTranFp(N-L), EnTranFp(L-G) and EnTranFp(N-G) show a small improvement compare to EnTranF in average RMSE.
	In panel (c) and (d) of Figure \ref{fig:RMSE Spread DoubleWell Penalty}, EnTranFp with linear transport map exhibits a similar ensemble spread  to EnKF. 
	But it has a significantly higher probability of covering the true state. EnTranFp with a nonlinear transport map reduces the Spread while simultaneously increasing the probability of covering the true state.
	Compare with the results of EnTranF in Figure \ref{fig:Spread DoubleWell}, the EnTranFp enhances the robustness of the EnTranF and leads to a better approximation of the posterior. {Finally, we directly compared the errors in the posterior variances estimated by EnTranF and EnKF against the posterior variance estimated by a particle filter with 10,000 particles used as a reference, as shown in Figure \ref{fig:cov_error_DW}. EnTranFp (L-G) and EnTranFp (N-G) get better accurate variance estimates compared to EnKF, while EnTranFp with linear kernel get a poor estimate of variance. }
	
	\subsection{Lorenz'63 System}\label{sec:4.3}
	Lorenz63 is a classic, chaotic system model proposed by American meteorologist Edward Lorenz in 1963. It describes the nonlinear dynamical behavior in atmospheric circulation. The Lorenz63 model is based on three partial differential equations and has three state variables: $X_1$, $X_2$, and $X_3$. This model is known for its highly sensitive initial conditions and nonlinear coupling. It leads to the famous ``butterfly effect''  where a small perturbation of initial { condition}  may result in long-term uncertainty and unpredictable outcomes. It can expressed in the form of ODEs
	\begin{equation}\label{eq:Lorenz63}
		\begin{aligned}
			\frac{\mathrm{d}X_1}{\mathrm{d}t} &= -\sigma X_1 + \sigma X_2 + \gamma_1\frac{\mathrm{d}B_1}{\mathrm{d}t},\\
			\frac{\mathrm{d}X_2}{\mathrm{d}t} &= -X_1X_3 + \rho X_1 - X_2 + \gamma_2\frac{\mathrm{d}B_2}{\mathrm{d}t},\\
			\frac{\mathrm{d}X_3}{\mathrm{d}t} &= X_1X_2 - \beta X_3 + \gamma_3\frac{\mathrm{d}B_3}{\mathrm{d}t},
		\end{aligned}
	\end{equation}
	where $B_1, B_2$ and $B_3$ are standard Brownian motion. Here, the parameters are set to be $\beta=8/3, \rho=28, \sigma=10$, which produces the well-known Lorenz attractor. {The noise level of the different components is set to be the same $\gamma_1=\gamma_2=\gamma_3=4\times 10^{-4}$.} We simulate the ODEs \eqref{eq:Lorenz63} by using a fourth-order Runge-Kutta method with time stepsize $\Delta t=0.01$. Here, we only take the observation operator to observe its first component. Partial observations greatly enhance the complexity and nonlinearity of the  filtering problem. At the same time, we consider observations with  Gaussian white noise. In this numerical example, we assimilate the Lorenz'63 system over $500$ { assimilation} windows. { A shallow DNN comprising $L=3$ layers with width $\{1, 40, 3\}$ and tanh activation functions is used to define the nudging term of the nonlinear transport map. The bandwidth of MMD was set to be the median of all $L_2$ distance between each pair of particles.}
	
\begin{figure}[!ht]
		\centering
		\subfigure{
			\begin{minipage}[t]{0.45\textwidth}
				\centering
				\includegraphics[width=\textwidth]{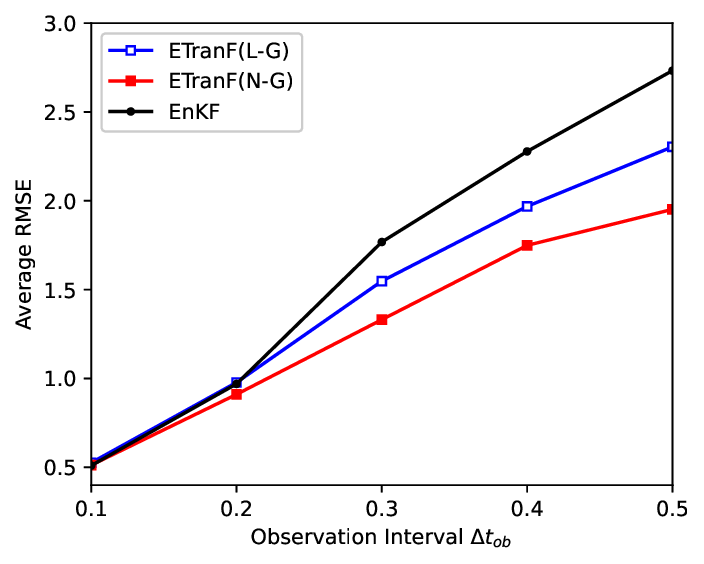}
			\end{minipage}
		}
		\subfigure{
			\begin{minipage}[t]{0.45\textwidth}
				\centering
				\includegraphics[width=\textwidth]{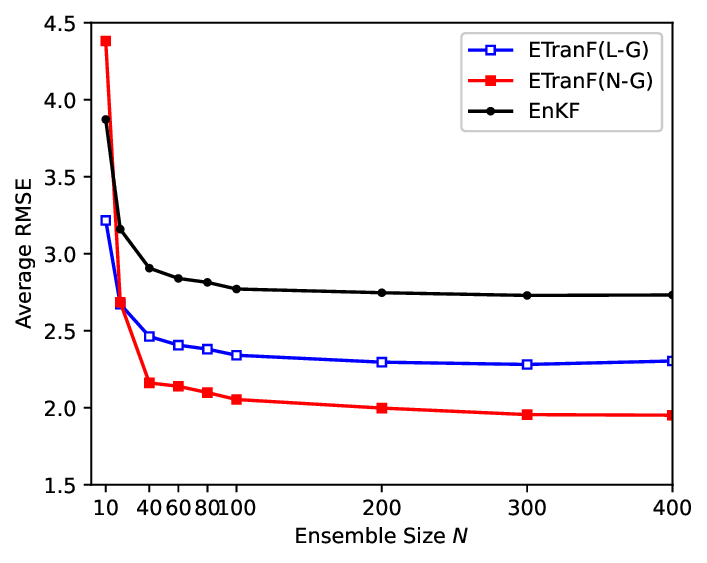}
			\end{minipage}
		}
		\caption{Average RMSE of Lorenz'63 system for ensemble size $N=400$ as a function of observation interval (Left), and for observation interval $\Delta t_{ob}=0.5$ as a function of ensemble size (Right).}
		\label{fig:rmse_lorenz63}
	\end{figure}

\begin{figure}[!h]
		\centering
		\subfigure{
			\begin{minipage}[t]{0.45\textwidth}
				\centering
				\includegraphics[width=\textwidth]{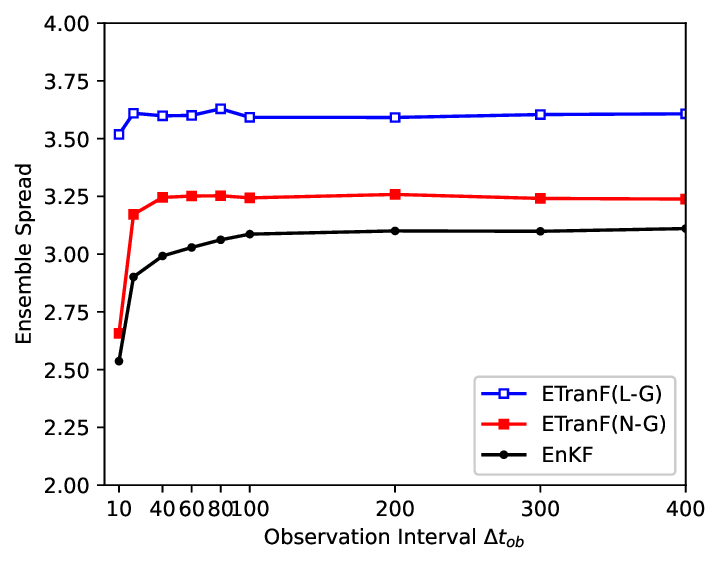}
			\end{minipage}
		}
		\subfigure{
			\begin{minipage}[t]{0.45\textwidth}
				\centering
				\includegraphics[width=\textwidth]{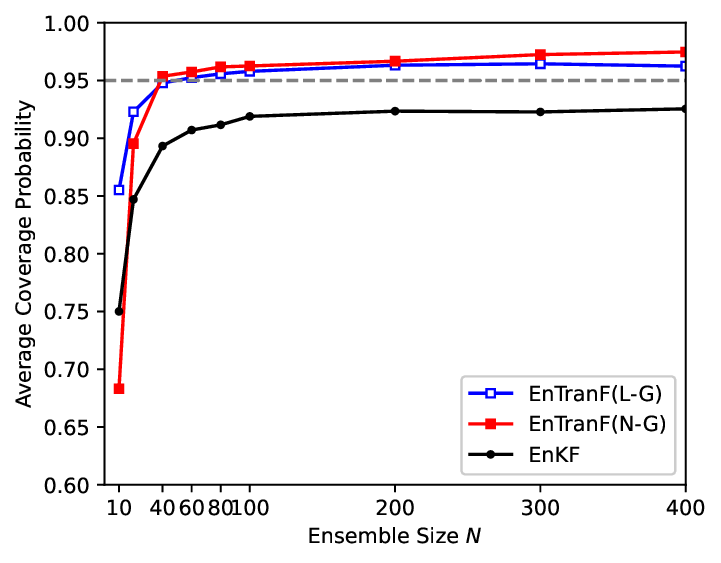}
			\end{minipage}
		}
		\caption{Spread of Lorenz'63 system for ensemble size $N=400$ as a function of observation interval (Left), and for observation interval $\Delta t_{ob}=0.5$ as a function of ensemble size (Right).}
		\label{fig:spread_lorenz63}
	\end{figure}

	At first,  we assimilate the Lorenz'63 system with the default setup of the ensemble transport filter. Figure \ref{fig:rmse_lorenz63} and Figure \ref{fig:spread_lorenz63} respectively show the 
	average RMSE and Spread versus  a range of observation interval and ensemble size.
	
	As shown in the left panel of Figure \ref{fig:rmse_lorenz63}, we demonstrate the average RMSE of EnTranF as a function of observation interval with a fixed ensemble size. 
	EnTranF(L-L) and EnTranF(N-L) experienced a collapse in the Lorenz'63 system, rendering it unable to produce meaningful state estimates. { This is because the linear kernel function results in significant information loss at each assimilation window, making the EnTranF ineffective in finding admissible transport maps as sequential filtering progresses.}
	EnTranF with a Gaussian kernel still performs well. EnTranF(L-G) and EnTranF(N-G) show a similar performance compared to EnKF when the observation interval is relatively short. For sufficiently large observation intervals, all variant of EnTranF present an increasing improvement on RMSE compared to EnKF as the observation interval increases. {EnTranF(N-G) and EnTranF(L-G) have $28.57\%$ and $15.89\%$ improvement on RMSE}, respectively, when taking observation interval as $\Delta t_{ob}=0.5$. A well-chosen kernel function can make EnTranF more robust in systems with strong nonlinearity.
	The right panel of Figure \ref{fig:rmse_lorenz63} shows the average RMSE of EnTranF as a function of ensemble size for the observation interval to be fixed. Since the EnTranF is based on Particle Filter, it exhibits almost no advantage over EnKF and may even perform worse when the ensemble size is small ($N<20$). However, EnTranF performs better than EnKF in the filtering problem of the Lorenz'63 system at larger ensemble sizes. {If ensemble size $N\ge 100$, EnTranF(L-G) has about $15.81\%$ improvement on average RMSE compared to EnKF, and EnTranF(N-G) has about $23.77\%$ improvement.} Unlike the one-dimensional double-well problem, the EnTranF with a nonlinear transport map has apparent advantages over the linear version. The relatively higher dimensional lorenz'63 system exhibits strong nonlinearities in partial observations, making methods based on linear transport maps less effective.
	
\begin{figure}[!h]
		\centering
		\subfigure{
			\begin{minipage}[t]{0.45\textwidth}
				\centering
				\includegraphics[width=\textwidth]{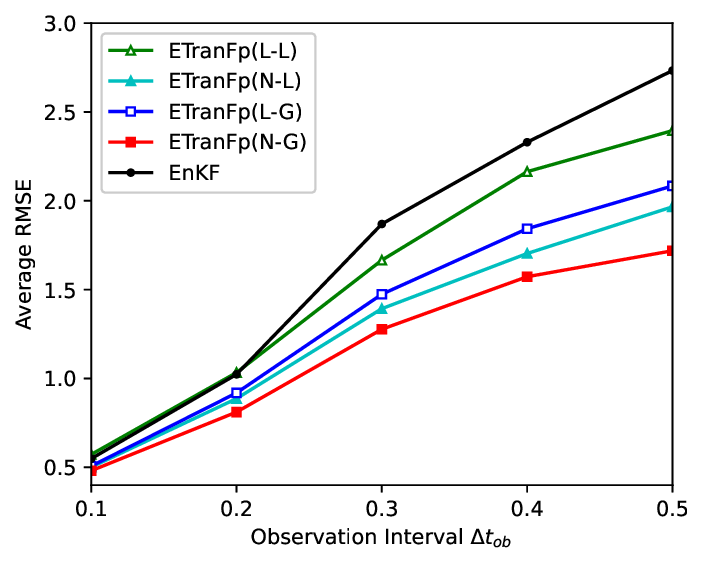}
			\end{minipage}
		}
		\subfigure{
			\begin{minipage}[t]{0.45\textwidth}
				\centering
				\includegraphics[width=\textwidth]{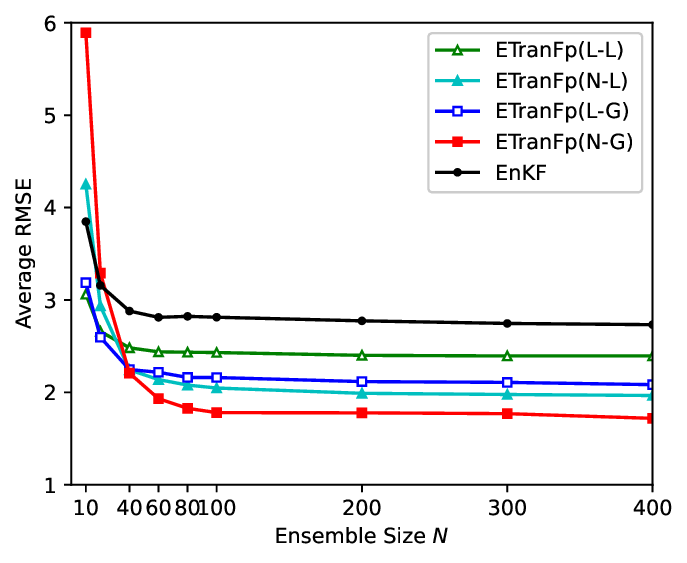}
			\end{minipage}
		}
		\caption{EnTranF with variance penalty: Average RMSE of Lorenz'63 system for ensemble size $N=400$ as a function of observation interval (Left), and for observation interval $\Delta t_{ob}=0.5$ as a function of ensemble size (Right).}
		\label{fig:rmse_lorenz63_penalty}
	\end{figure}

\begin{figure}[!htbp]
		\centering
		\subfigure{
			\begin{minipage}[t]{0.45\textwidth}
				\centering
				\includegraphics[width=\textwidth]{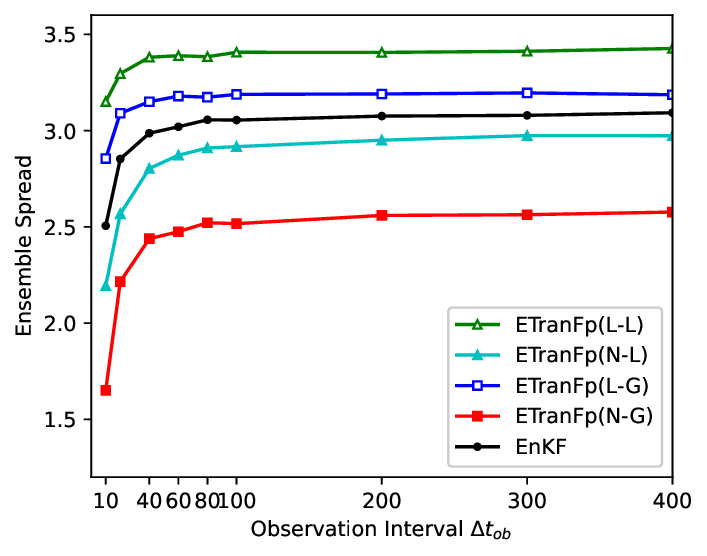}
			\end{minipage}
		}
		\subfigure{
			\begin{minipage}[t]{0.45\textwidth}
				\centering
				\includegraphics[width=\textwidth]{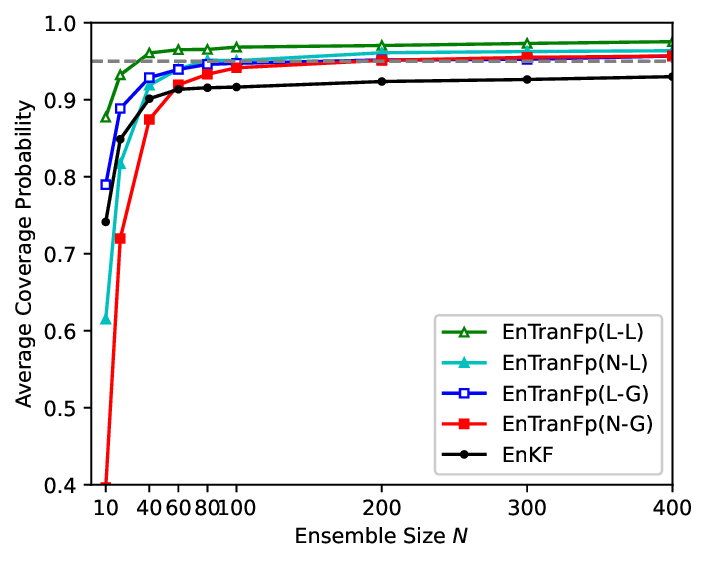}
			\end{minipage}
		}
		\caption{EnTranF with variance penalty: Spread of Lorenz'63 system for ensemble size $N=400$ as a function of observation interval (Left), and for observation interval $\Delta t_{ob}=0.5$ as a function of ensemble size (Right).}
		\label{fig:spread_lorenz63_penalty}
	\end{figure}

	The ensemble spread of different filtering methods is shown in Figure \ref{fig:spread_lorenz63}. EnTranF(N-G) reduces the ensemble spread compared to EnTranF(L-G). However, both EnTranF(N-G) and EnTranF(L-G) have a larger ensemble spread with different observation intervals and ensemble sizes. This may be caused by inadequate optimization of the MMD loss function.   EnTranF fails to approximate the posterior adequately. The EnTranF with a variance penalty can significantly improve the  spread and slightly improve the average RMSE. Figure \ref{fig:rmse_lorenz63_penalty} and Figure \ref{fig:spread_lorenz63_penalty} respectively depict  the average RMSE and Spread over a range of observation intervals and ensemble sizes.
	
	The average RMSE of EnTranF with a variance penalty outperforms EnKF at different observation intervals  (see left panel of Figure \ref{fig:rmse_lorenz63_penalty}). After introducing the variance penalty term, even EnTranFp with a linear kernel function shows a significant performance improvement on average RMSE compared to EnKF. {EnTranF(L-L) has  $12.37\%$ improvement, and EnTranF(N-L) has  $28.06\%$ improvement. EnTranFp also shows some improvement compared to EnTranF when taking a Gaussian kernel function. Here, EnTranFp(L-G) is $23.76\%$ better than EnKF on RMSE at observation interval $\Delta t_{ob}=0.5$, while EnTranFp(N-G) is $37.11\%$ better than EnKF.} From right panel of Figure \ref{fig:rmse_lorenz63_penalty}, EnTranFp has a smaller average RMSE when $N\ge 20$ than EnKF for different ensemble sizes. In Figure \ref{fig:spread_lorenz63_penalty}, EnTranFp(L-L), EnTranFp(N-L), and EnTranFp(L-G) show a similar spread to  EnKF but get a higher probability of covering the true state. EnTranFp(N-G) performs optimally in the Lorenz63 system. It significantly reduces the spread while obtaining a probability of covering the true state close to $95\%$. The variance penalty term effectively guides EnTranF in prioritizing the approximation of high-informative statistics of the posterior distribution. It enhances the robustness of EnTranF in high-dimensional nonlinear systems.

	\subsection{Lorenz'96 System}\label{sec:4.4}
	The Lorenz'96 model is a simplified mathematical model that describes the nonlinear dynamical behavior in atmospheric circulation systems. It consists of a set of coupled one-dimensional differential equations, each representing a grid point within the atmospheric circulation system. These differential equations are typically given by:
	\begin{equation}\label{eq:lorenz96}
		\frac{{\mathrm{d}X_i}}{{\mathrm{d}t}} = (X_{i+1} - X_{i-2})X_{i-1} - X_i + F + \gamma_i\frac{\mathrm{d}B_i}{\mathrm{d}t}.
	\end{equation}
	Here, $X_i$ represents the state variable at the $i$-th grid point, and $F$ is an external forcing. {The noise level of the different components is set to be the same $\gamma_i=4\times 10^{-4}$.}
	
\begin{figure}[!h]
		\centering
		\includegraphics[width=\textwidth]{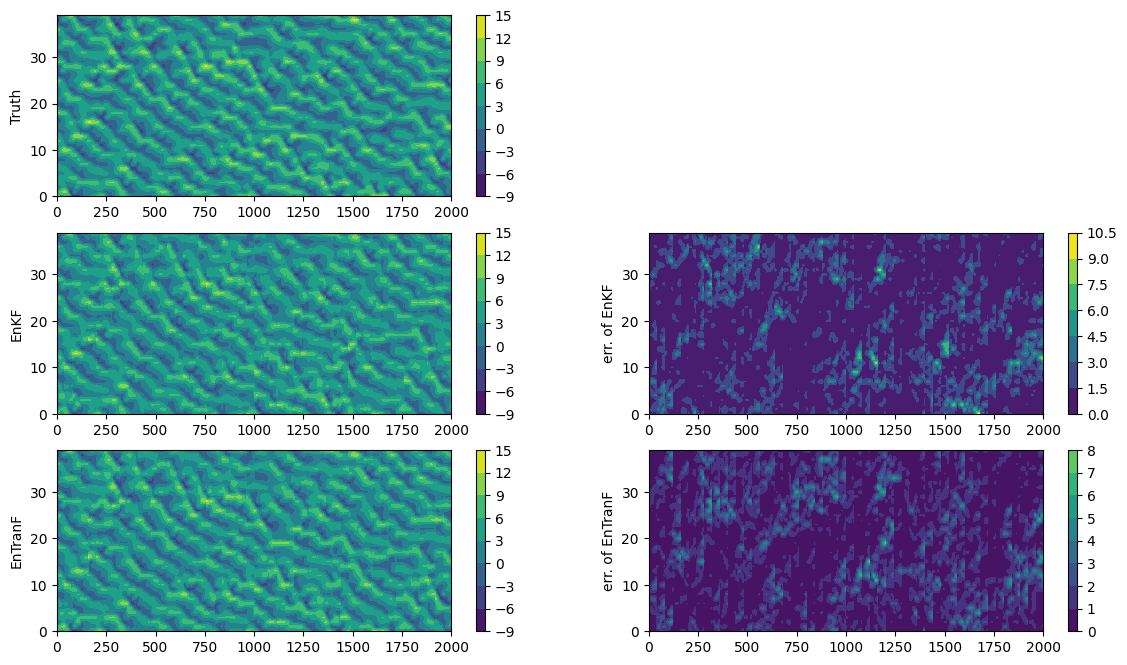}
		\caption{State estimation and absolute error of EnKF and EnTranF for Lorenz'96 system with observation interval $\Delta t_{ob}=0.4$ and ensemble size $N=4000$.}
		\label{fig:lorenz96_err}
	\end{figure}

	The numerical example is  conducted to evaluate the performance of EnTranF for a of Lorenz'96 system in which sparse observations are given both in space and time. We take 40-dimension state variables and observe 20 dimensions of them  (observing every other state component). We simulate the ODEs \eqref{eq:lorenz96} by using a fourth-order Runge-Kutta method with time stepsize $\Delta t=0.01$. The observation interval in time was set to be $\Delta_{ob} = 0.4$ and {assimilation steps was set to be $500$.} The observation noise is set to be white noise with a variance of $1.0$. Numerical results are presented in Figure \ref{fig:lorenz96_err} and Figure \ref{fig:metric_lorenz96}. To address the Lorenz'96 system, we split the 20-dimensional observation vector into components, each consisting of two dimensions. Then, prior ensembles are sequentially transform to posterior ensembles using the approach detailed in Section \ref{sec:3.2.2}, which allows for more effective ensemble transport filter in this high-dimensional setting. { For each component, the transport map was defined using a DNN comprising $L=4$ layers with widths $\{2, 10, 10, 40\}$ and tanh activation functions to parametric the nudging term. The bandwidth of MMD was set to be the median of all $L_2$ distance between each pair of particles.}
	
	{Figure \ref{fig:lorenz96_err} shows partial states over time-horizen of Lorenz'96 estimated using EnKF and EnTranF(N-G), respectively, along with their absolute errors relative to the true state. In this 40-dimensional problem, the PF cannot be implemented with limited ensemble size due to the particle degeneracy issue. EnTranF(N-G) obtained a correct state estimation of the Lorenz'96 state, with a similar absolute error compare to EnKF. The results show that EnTranF, which reconstructs the resampling step via transport maps, is able to rescure the issue of particle degeneracy.
		
		\begin{figure}[!htbp]
			\centering
			\includegraphics[width=.5\textwidth]{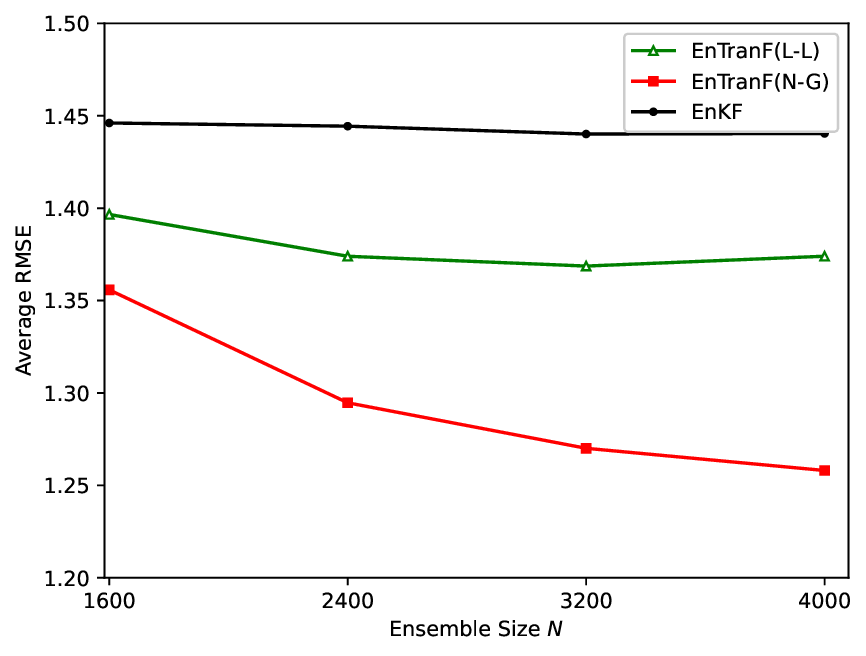}
			\caption{Average RMSE of EnKF, EnTranF(L-L) and EnTranF(N-G) for Lorenz'96 system with observation interval $\Delta t_{ob}=0.4$ as a function of ensemble size.}
			\label{fig:metric_lorenz96}
		\end{figure}
		
		The average RMSE of EnKF, EnTranF(L-L) and EnTranF(N-G) are shown in Figure \ref{fig:metric_lorenz96}. It presents that both EnTranF(L-L) and EnTranF(N-G) exhibit superior performance compared to the EnKF, and as the number of particles increases, the performance of EnTranF(N-G) gradually improves. When the ensemble size exceeds 1600, EnTranF achieves an average RMSE reduction of 10.34\% compared to EnKF. Here, EnTranF requires a large ensemble size compare to EnKF, primarily because it relies on sampling particles to approximate these distributions and does not perform state space localization. This means that to accurately capture the characteristics of the target distribution, a large number of particles is usually required. When the ensemble size is small, the inadequacy in the number of particles can lead to insufficient accuracy in the approximation of the distribution, thereby affecting the effectiveness of the algorithm.
		
		For smaller ensemble sizes, we compare EnTranF with LETKF both in a domain localization setting with localization radius $r_{loc}=2$. The inflation parameter for the LETKF is set to be $1.2$. In this setup, the dimension of each local block is 5 and the data assimilation for each local block can be performed in parallel. The average RMSE and coverage probability of EnKF and EnTranF are shown in Figure \ref{fig:metric_L96_Domain_loc}. 
		\begin{figure}[!h]
			\centering
			\subfigure{
				\begin{minipage}[t]{0.45\textwidth}
					\centering
					\includegraphics[width=\textwidth]{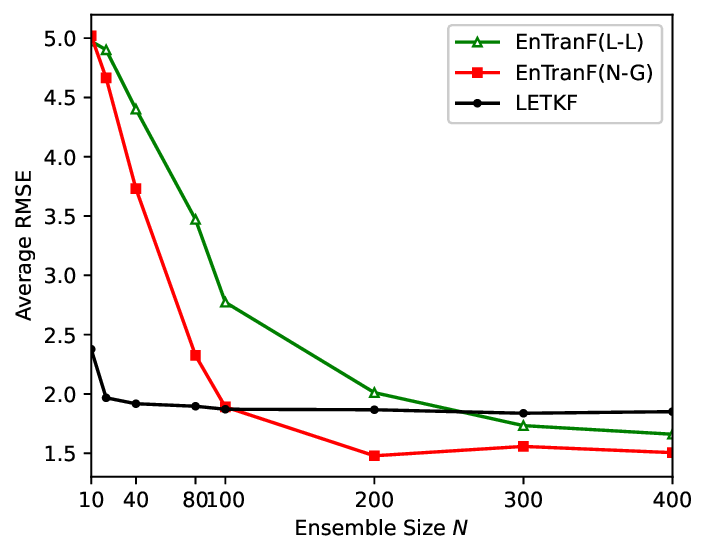}
				\end{minipage}
			}
			\subfigure{
				\begin{minipage}[t]{0.45\textwidth}
					\centering
					\includegraphics[width=\textwidth]{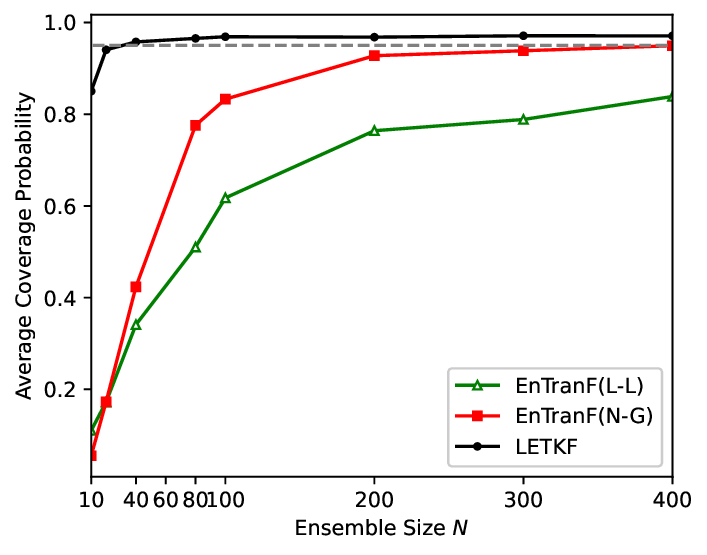}
				\end{minipage}
			}
			\caption{Average RMSE and Coverage probability of LETKF and EnTranF for Lorenz'96 system with observation interval $\Delta t_{ob}=0.4$ as a function of ensemble size.}
			\label{fig:metric_L96_Domain_loc}
		\end{figure}
		
		From the left panel of Figure \ref{fig:metric_L96_Domain_loc}, EnTranF(N-G) achieves a lower RMSE than LETKF when the ensemble size $N$ exceeds $100$, resulting in more accurate state estimation. EnTranF(L-L) requires an ensemble size exceeding $300$ to achieve performance slightly better than that of LETKF. EnTranF(L-L) can achieve a coverage rate close to $95\%$ when the ensemble size exceeds 200, indicating accurate uncertainty estimation. EnTranF(L-L) performs generally poorly in the high-dimensional Lorenz'96 system. The linear kernel and linear map make it difficult to adequately capture the features of complex high-dimensional distributions, thereby reducing its effectiveness in sequential data assimilation tasks. 
	}
	
	\section{Conlusion}\label{sec:5}
	In this paper, we reconstructed the analysis step of the particle filter through a transport map, which directly transported the prior particles to the posterior particles. By matching the expectation information of posterior distribution, the problem was converted into an optimization problem for the Maximum Mean Discrepancy loss function. 
	In the ensemble transport filter, the particle weights in the analysis step of the particle filter were substituted with the update  of particle states, effectively alleviating the risk of particle degeneracy in high-dimensional problems. Meanwhile, the ensemble transport filter inherited the accurate estimation of the posterior distribution from the particle filter. {To improve the robustness of MMD, we introduced a variance penalty term to guide the prioritized optimization of high-informative statistics in reference posterior. }
	
	Numerical results  demonstrated that the ensemble transport filter outperformed EnKF significantly in lower-dimensional problems, particularly in strongly nonlinear systems. The ensemble transport filter remained efficient in high-dimensional systems and performed comparably to EnKF. The variance penalty term effectively enhanced the robustness of our method in approximating the filtering posterior, making it more effective and stable. EnTranF with a variance penalty term significantly reduced the spread of ensemble methods while maintaining RMSE. 
	
	The general transport map structure can be explored for future work to avoid computationally expensive  optimization problems when constructing the transport map at each assimilation window. {Furthermore, Motivated by the block-triangular structure of transport map \cite{pmlr-v235-al-jarrah24a}, a likelihood-free version of our mehtod can be proposed that avoid a optimization on a min-max problem. }
	
	\smallskip
	\bigskip
	\textbf{Acknowledgement:}
	L. Jiang acknowledges the support of NSFC 12271408.


	\newpage
	
	\section{Appendix}
	
		\subsection{Toy Example for the Efficacy of Variance Penalty}
\begin{figure}[!htbp]
			\centering
			\includegraphics[width=.85\textwidth]{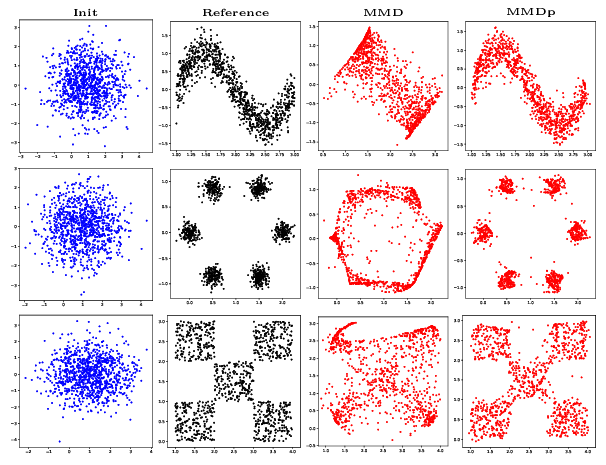}
			\caption{Tree toy example for the comparision of MMD and MMDp. The blue points were sampled from a two dimensional Gaussian distribution (Init), and the black points were generated by three complicated distributions (Reference). Red ponints presented the approximated distribution by minimizing the MMD and MMDp repectively.}
			\label{fig:efficacy of MMDp}
		\end{figure}

		In this subsection, we present toy examples on three synthetic datasets to demonstrate the effectiveness of introducing the variance penalty term \eqref{eq:penalty MMD} in the MMD framework. As shown in Figure \ref{fig:efficacy of MMDp}, three sets of sample points were generated from multimodal distributions defined on a two-dimensional space. In this experiment, we start with a two-dimensional Gaussian distribution and adjust the transport map by minimizing the MMD and MMDp, respectively, to transform the Gaussian distribution into the reference distribution. For all three loss functions, we use a Gaussian kernel with a bandwidth of $b_w=0.05$. A fully connected shallow neural network with hidden layer widths of $[10, 10]$ is employed to parameterize the transport map.
		
		The first row of Figure \ref{fig:efficacy of MMDp} shows a relatively simple sine-shaped distribution. The approximate distribution obtained using MMD as the loss function captures the overall shape of the reference distribution, but it fails to accurately reproduce the spread in the central region. In contrast, when using MMDp as the loss function, the resulting approximate distribution provides a much more accurate representation of the reference distribution. The second and third rows of Figure \ref{fig:efficacy of MMDp} present two multimodal distributions. The approximate distributions obtained using MMD as the loss function provide a rough match to the overall shape, but fail to accurately separate the individual modes. In contrast, MMDp, which incorporates a variance penalty, is able to capture the multimodal structure of the reference distributions more accurately. This indicates that the variance penalty can effectively guide the approximate distribution to learn the fine structure of the reference distribution.

		\subsection{Example of Lorenz'63 with observing the third component $X_3$}
		
\begin{figure}[!htbp]
			\centering
			\subfigure[EnTranF]{
				\begin{minipage}[t]{0.45\textwidth}
					\centering
					\includegraphics[width=\textwidth]{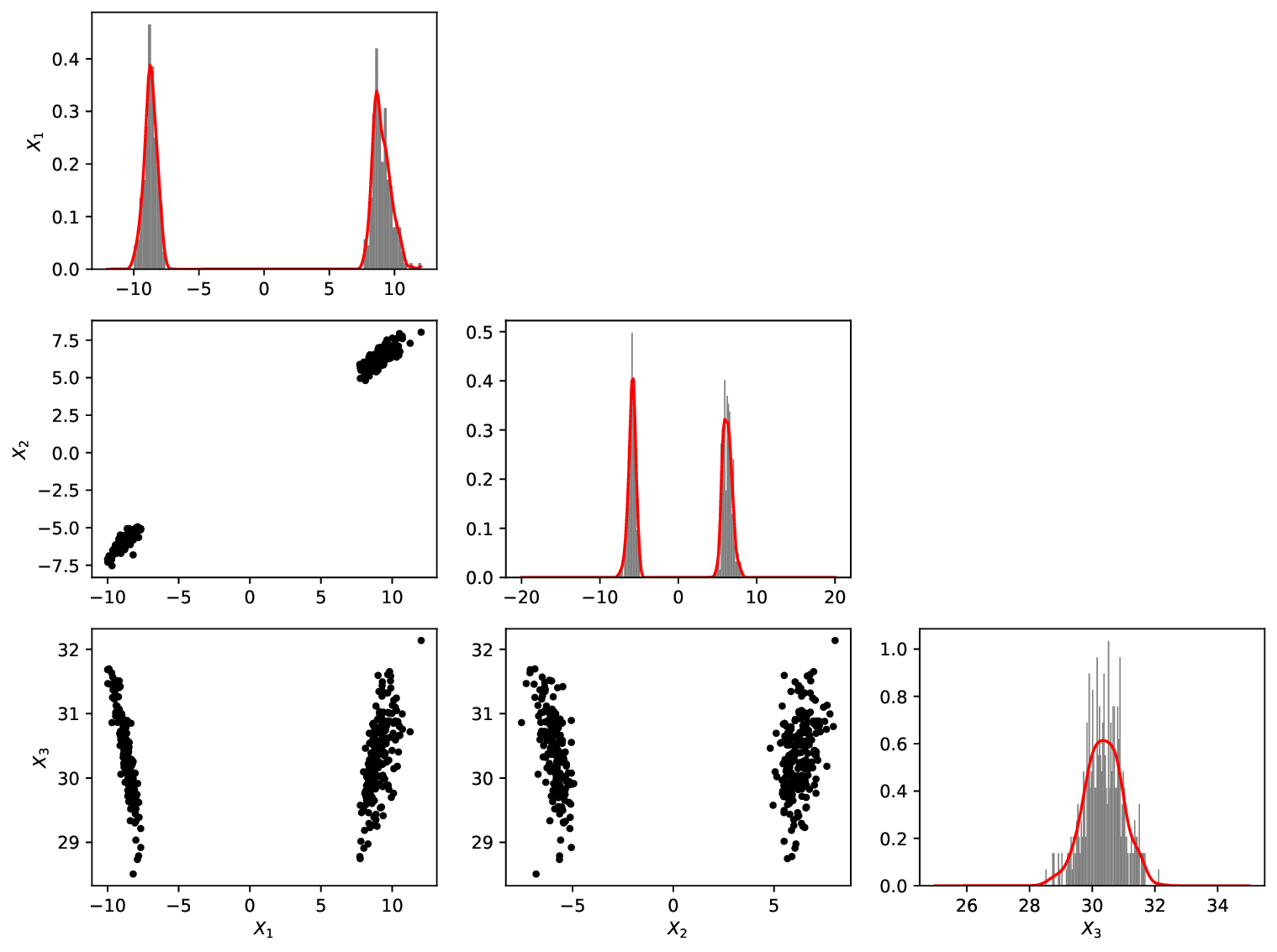}
				\end{minipage}
			}
			\subfigure[EnKF]{
				\begin{minipage}[t]{0.45\textwidth}
					\centering
					\includegraphics[width=\textwidth]{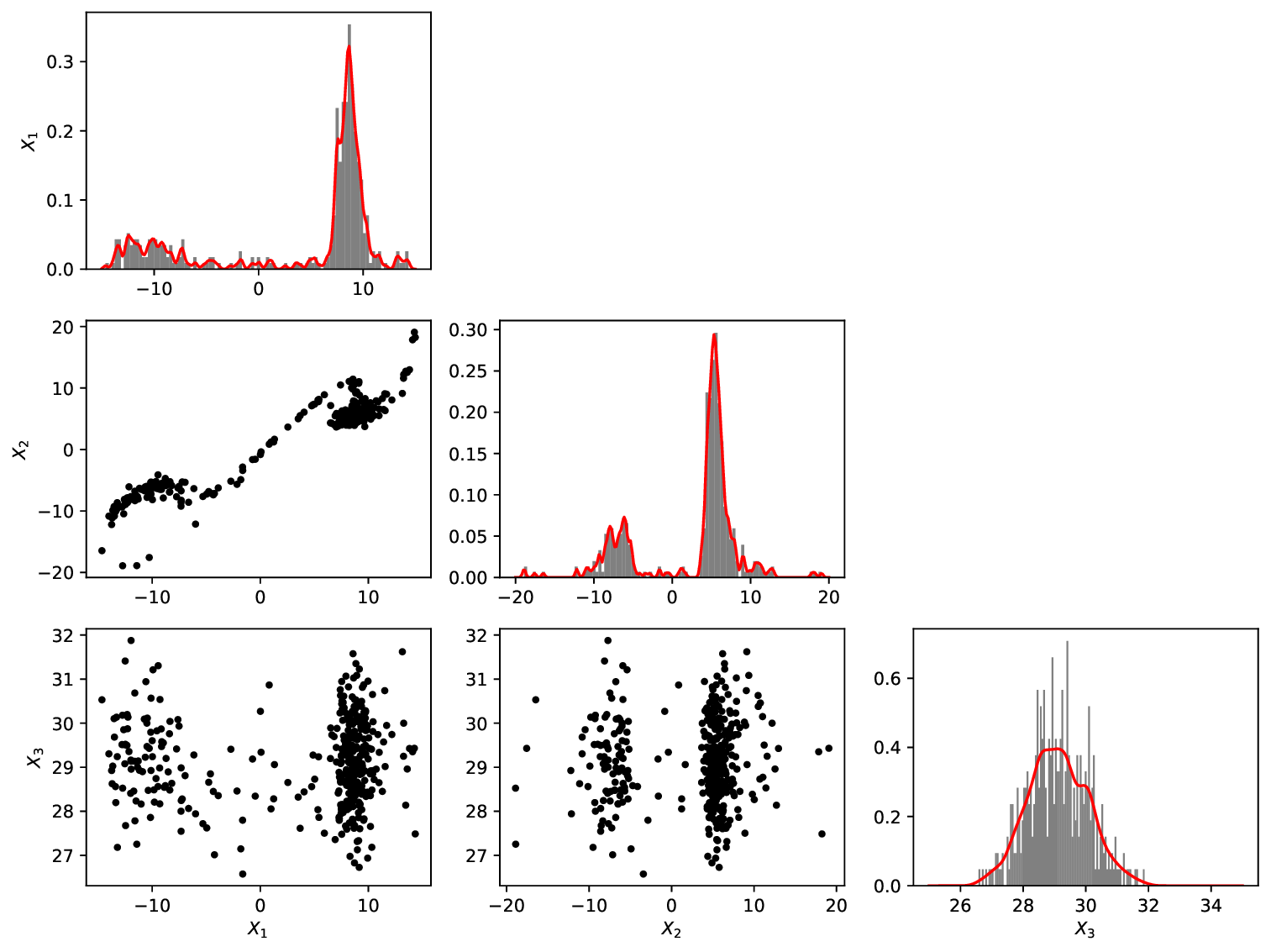}
				\end{minipage}
			}
			\caption{One-dimensional and two-dimensional marginal posterior distributions of posterior generated by EnTranF and EnKF at $t=5.0$}
			\label{fig:marginal_distro_L63_X3}
		\end{figure}

		In data assimilation, if only the third component $X_3$ of the Lorenz '63 system is observed, the corresponding posterior distribution exhibits a multimodal nature, because $X_3$ cannot distinguish whether the system is evolving on the left or the right lobe of the attractor. Here, we adopt the same problem and parameter settings as in Section 4.3 for the Lorenz'63 system. Due to the multi-modal nature of the posterior distribution, using the mean as the state estimate for filtering methods is inaccurate. Here, Figure \ref{fig:marginal_distro_L63_X3} show scatters of the particle ensemble at time $t=5.0$ to illustrate the effectiveness of different filtering methods in estimating the posterior distribution.
		
		From Figure \ref{fig:marginal_distro_L63_X3}, both EnKF and EnTranF exhibit multi-modal marginal posterior distributions for states X1 and X2, while X3 is approximately Gaussian. The posterior distribution estimated by EnTranF assigns nearly equal probabilities to the two attractors, whereas EnKF shows a stronger bias towards one attractor.


\begin{thebibliography}{99}
		
		\bibitem{pmlr-v235-al-jarrah24a} 
		{\sc M. Al-Jarrah, N. Jin, B. Hosseini and A. Taghvaei}, {\em Nonlinear filtering with brenier optimal transport maps}, PMLR, 235 (2024), pp.813--839.
		
		\bibitem{brajardCombiningDataAssimilation2021} 
		{\sc J. Brajard, A. Carrassi, M. Bocquet and L. Bertino}, {\em Combining data assimilation and machine learning to infer unresolved scale parametrization}, Philos. Trans. R. Soc. A, 379 (2021), p.20200086.{https://doi.org/10.1098/rsta.2020.0086}.
		
		\bibitem{carrassiDataAssimilationGeosciences2018} 
		{\sc A. Carrassi, M. Bocquet, L. Bertino and G. Evensen}, {\em Data assimilation in the geosciences: An overview of methods, issues, and perspectives}, WIREs Clim. Change, 9 (2018), p.e535.{https://doi.org/10.1002/wcc.535}.
		
		\bibitem{chattopadhyayDeepLearningenhancedEnsemblebased2023} 
		{\sc A. Chattopadhyay, E. Nabizadeh, E. Bach and P. Hassanzadeh}, {\em Deep learning-enhanced ensemble-based data assimilation for high-dimensional nonlinear dynamical systems}, J. Comput. Phys., 477 (2023), p.111918.{https://doi.org/10.1016/j.jcp.2023.111918}.
		
		\bibitem{elmoselhyBayesianInferenceOptimal2012} 
		{\sc T. A. El Moselhy, and Y. M. Marzouk}, {\em Bayesian inference with optimal maps}, J. Comput. Phys., 231 (2012), pp.7815--7850.{https://doi.org/10.1016/j.jcp.2012.07.022}.
		
		\bibitem{evensenEnsembleKalmanFilter2003} 
		{\sc G. Evensen}, {\em The ensemble kalman filter: Theoretical formulation and practical implementation}, Ocean Dynam., 53 (2003), pp.343--367.{https://doi.org/10.1007/s10236-003-0036-9}.
		
		\bibitem{evensenSamplingStrategiesSquare2004} 
		{\sc G. Evensen}, {\em Sampling strategies and square root analysis schemes for the EnKF}, Ocean Dynam., 54 (2004), pp.539--560.{https://doi.org/10.1007/s10236-004-0099-2}.
		
		\bibitem{evensenEnsembleKalmanFilter2009} 
		{\sc G. Evensen}, {\em The ensemble Kalman filter for combined state and parameter estimation}, IEEE Control Syst., 29 (2009), pp.83--104.{https://doi.org/10.1109/MCS.2009.932223}.
		
		\bibitem{farchiReviewArticleComparison2018} 
		{\sc A. Farchi and M. Bocquet}, {\em Review article: Comparison of local particle filters and new implementations}, Nonlinear Process. Geophys., 25 (2018), pp.765--807.{https://doi.org/10.5194/npg-25-765-2018}.
		
		\bibitem{farchiUsingMachineLearning2021} 
		{\sc A. Farchi, P. Laloyaux, M. Bonavita and M. Bocquet}, {\em Using machine learning to correct model error in data assimilation and forecast applications}, Q. J. R. Meteorol. Soc., 147 (2021), pp.3067--3084.{https://doi.org/10.1002/qj.4116}.
		
		\bibitem{NIPS2006_e9fb2eda} 
		{\sc A. Gretton, K. Borgwardt, M. Rasch, B. Sch\"{o}lkopf and A. Smola}, {\em A kernel method for the two-sample-problem}, NeurIPS, 19 (2006), pp.513--520. 
		
		\bibitem{higham.AlgorithmicIntroductionNumerical2001} 
		{\sc D. J. Higham}, {\em An algorithmic introduction to numerical simulation of stochastic differential equations}, SIAM Rev., 43 (2001), pp.525--546.{https://doi.org/10.1137/S0036144500378302}.
		
		\bibitem{hoangMachineLearningbasedConditional2023} 
		{\sc T.-V. Hoang, S. Krumscheid, H.G. Matthies and R. Tempone}, {\em Machine learning-based conditional mean filter: A generalization of the ensemble kalman filter for nonlinear data assimilation}, Found. Data Sci., 5 (2023), pp.56--80.{https://doi.org/10.3934/fods.2022016}.
		
		\bibitem{jiangCorrectingNoisyDynamic2022} 
		{\sc L. Jiang and N. Liu}, {\em Correcting noisy dynamic mode decomposition with Kalman filters}, J. Comput. Phys., 461 (2022), p.111175.{https://doi.org/10.1016/j.jcp.2022.111175}.
		
		\bibitem{kawabataNonGaussianProbabilityDensities2020}
		{\sc T. Kawabata and G. Ueno}, {\em Non-Gaussian probability densities of convection initiation and development investigated using a particle filter with a storm-scale numerical weather prediction model}, Mon. Weather Rev., 148 (2020), pp.3--20.{https://doi.org/10.1175/MWR-D-18-0367.1}.
		
		\bibitem{lawDataAssimilationMathematical2015}
		{\sc K. Law, A. Stuart, and K. Zygalakis}, {\em Data Assimilation: A Mathematical Introduction}, Springer, Cham, 2015.{https://doi.org/10.1007/978-3-319-20325-6}.
		
		\bibitem{liptserStatisticsRandomProcesses1977}
		{\sc R. S. Liptser and A. N. Shiryayev}, {\em Statistics of Random Processes I: General Theory}, Springer, New York, 1977.{https://doi.org/10.1007/978-1-4757-1665-8}.
		
		\bibitem{liptserStatisticsRandomProcesses2001}
		{\sc R. S. Liptser and A. N. Shiryaev}, {\em Statistics of Random Processes II: Applications}, Springer, Heidelberg, 2001.{https://doi.org/10.1007/978-3-662-10028-8}.
		
		\bibitem{liuPerronFrobeniusOperator2024}
		{\sc N. Liu and L. Jiang}, {\em Perron--frobenius operator filter for stochastic dynamical systems}, SIAM/ASA J. Uncertain. Quantif., 12 (2024), pp.182--211.{https://doi.org/10.1137/23M1547391}.
		
		\bibitem{majdaFilteringComplexTurbulent2012}
		{\sc A. J. Majda and J. Harlim}, {\em Filtering Complex Turbulent Systems}, Cambridge University Press, 2012.{https://doi.org/10.1017/CBO9781139061308}.
		
		\bibitem{mandelConvergenceEnsembleKalman2011}
		{\sc J. Mandel, L. Cobb, and J. D. Beezley}, {\em On the convergence of the ensemble kalman filter}, Appl. Mat., 56 (2011), pp.533--541.{https://doi.org/10.1007/s10492-011-0031-2}.
		
		\bibitem{mongeMemoireTheorieDeblais1781}
		{\sc G. Monge}, {\em M{\'e}moire sur la th{\'e}orie des d{\'e}blais et des remblais}, Mem. Math. Phys. Acad. Royale Sci., (1781), pp. 666--704.
		
		\bibitem{pulidoSequentialMonteCarlo2019}
		{\sc M. Pulido and P. J. Van Leeuwen}, {\em Sequential monte carlo with kernel embedded mappings: The mapping particle filter}, J. Comput. Phys., 396 (2019), pp.400--415.{https://doi.org/10.1016/j.jcp.2019.06.060}.
		
		\bibitem{ramdasDecreasingPowerKernel2015}
		{\sc A. Ramdas, S. J. Reddi, B. Poczos, A. Singh and L. Wasserman}, {\em On the decreasing power of kernel and distance based nonparametric hypothesis tests in high dimensions}, AAAI, 29 (2015).
		
		\bibitem{rebeschiniCanLocalParticle2015}
		{\sc P. Rebeschini and R. Van Handel}, {\em Can local particle filters beat the curse of dimensionality?}, The Annals of Applied Probability, 25 (2015).
		
		\bibitem{richardmRealAnalysisProbability2018}
		{\sc D. Richard M}, {\em Real Analysis and Probability}, {Chapman and Hall/CRC}, 2002.
		
		\bibitem{sakovIterativeEnKFStrongly2012}
		{\sc P.~Sakov, D.~S. Oliver, and L.~Bertino}, {\em An iterative EnKF for strongly nonlinear systems}, Mon. Weather Rev., 140 (2012), pp.1988--2004.
		
		\bibitem{snyderObstaclesHighdimensionalParticle2008}
		{\sc C.~Snyder, T.~Bengtsson, P.~Bickel, and J.~Anderson}, {\em Obstacles to high-dimensional particle filtering}, Mon. Weather Rev., 136 (2008), pp.~4629--4640.
		
		\bibitem{spantiniCouplingTechniquesNonlinear2022}
		{\sc A.~Spantini, R.~Baptista, and Y.~Marzouk}, {\em Coupling techniques for nonlinear ensemble filtering}, SIAM Rev., 64 (2022), pp.~921--953.
		
		\bibitem{vanleeuwenParticleFilteringGeophysical2009}
		{\sc P.~J. Van~Leeuwen}, {\em Particle filtering in geophysical systems}, Mon. Weather Rev., 137 (2009), pp.~4089--4114.
		
		\bibitem{vanleeuwenNonlinearDataAssimilation2015}
		{\sc P.~J. Van~Leeuwen, Y.~Cheng, and S.~Reich}, {\em Nonlinear Data Assimilation}, Springer, Cham, 2015.
		
		\bibitem{villaniOptimalTransportOld2009}
		{\sc C.~Villani}, {\em Optimal Transport: Old and New}, Springer, Heidelberg, 2009.
		
		\bibitem{wangPhysicsInformedDeep2022}
		{\sc Z.~Wang, W.~Xing, R.~Kirby, and S.~Zhe}, {\em Physics informed deep kernel learning}, PMLR, 151 (2022), pp.~1206--1218.
		
		\bibitem{wangEfficientNeuralNetwork2025}
		{\sc Z.~O. Wang, R.~Baptista, Y.~Marzouk, L.~Ruthotto, and D.~Verma}, {\em Efficient neural network approaches for conditional optimal transport with applications in bayesian inference}, SIAM J. Sci. Comput., 47 (2025), pp.C979-C1005.
		
		\bibitem{yanWeightedClassSpecificMaximum2020}
		{\sc H.~Yan, Z.~Li, Q.~Wang, P.~Li, Y.~Xu, and W.~Zuo}, {\em Weighted and class-specific maximum mean discrepancy for unsupervised domain adaptation}, IEEE Trans. Multimedia, 22 (2020), pp.~2420--2433.
		
		\bibitem{zhangCoupledDataAssimilation2020}
		{\sc S.~Zhang, Z.~Liu, X.~Zhang, X.~Wu, G.~Han, Y.~Zhao, X.~Yu, C.~Liu, Y.~Liu, S.~Wu, F.~Lu, M.~Li, and X.~Deng}, {\em Coupled data assimilation and parameter estimation in coupled ocean--atmosphere models: A review}, Clim. Dyn., 54 (2020), pp.~5127--5144.
		
		\bibitem{zhangMaximumMeanCovariance2020}
		{\sc W.~Zhang, X.~Zhang, L.~Lan, and Z.~Luo}, {\em Maximum mean and covariance discrepancy for unsupervised domain adaptation}, Neural Process. Lett., 51 (2020), pp.~347--366.
		
		\bibitem{zhuImplicitEqualWeights2016}
		{\sc M.~Zhu, P.~J. Van~Leeuwen, and J.~Amezcua}, {\em Implicit equal-weights particle filter}, Q. J. R. Meteorol. Soc., 142 (2016), pp.~1904--1919.
		
		\bibitem{houtekamerSequentialEnsembleKalman2001}
		{\sc P. L. Houtekamer and H. L. Mitchell}, {\em A Sequential Ensemble Kalman Filter for Atmospheric Data Assimilation}, Mon. Weather Rev., 129(2001), pp.123--137.
		
		\bibitem{andersonOptimalFiltering1979}
		{\sc B. D. O. Anderson and J. B. Moore}, {\em Optimal Filtering}, Prentice-Hall, Englewood Cliffs, 1979.
		
	\end{thebibliography}
\end{document}